\documentclass[sigconf]{acmart}

\usepackage{graphicx} 

\usepackage{amsmath,amsfonts} 
\usepackage{textcomp}
\usepackage{xcolor}
\usepackage{bm}
\usepackage{hyperref}
\usepackage{amsthm}
\usepackage{subfigure}
\usepackage[linesnumbered,ruled,noend]{algorithm2e} 
\usepackage{algorithmic}
\usepackage{multirow}
\usepackage{float}
\usepackage{siunitx}

\AtBeginDocument{%
  \providecommand\BibTeX{{%
    \normalfont B\kern-0.5em{\scshape i\kern-0.25em b}\kern-0.8em\TeX}}}

\copyrightyear{2022}
\acmYear{2022}
\setcopyright{acmcopyright}

\acmConference[WWW '22]{Proceedings of the ACM Web Conference 2022}{April 25--29, 2022}{Virtual Event, Lyon, France}
\acmBooktitle{Proceedings of the ACM Web Conference 2022 (WWW '22), April 25--29, 2022, Virtual Event, Lyon, France}
\acmPrice{15.00}
\acmISBN{978-1-4503-9096-5/22/04}
\acmDOI{10.1145/3485447.3512030}

\begin{document}
\title{EXIT: Extrapolation and Interpolation-based Neural Controlled\\Differential Equations for Time-series Classification and Forecasting}

\author{Sheo Yon Jhin, Jaehoon Lee, Minju Jo, Seungji Kook, Jinsung Jeon, Jihyeon Hyeong, Jayoung Kim, and Noseong Park}
\affiliation{%
  \institution{Yonsei University}
  \city{Seoul}
  \country{South Korea}}
\email{{sheoyonj,ljh5694,alflsowl12,2021321393,jjsjjs0902,jiji.hyeong,jayoung.kim,noseong}@yonsei.ac.kr}








\renewcommand{\shortauthors}{Jhin et al.}

\begin{abstract}
Deep learning inspired by differential equations is a recent research trend and has marked the state of the art performance for many machine learning tasks. Among them, time-series modeling with neural controlled differential equations (NCDEs) is considered as a breakthrough. In many cases, NCDE-based models not only provide better accuracy than recurrent neural networks (RNNs) but also make it possible to process irregular time-series. In this work, we enhance NCDEs by redesigning their core part, i.e., generating a continuous path from a discrete time-series input. NCDEs typically use interpolation algorithms to convert discrete time-series samples to continuous paths. However, we propose to i) generate another latent continuous path using an encoder-decoder architecture, which corresponds to the interpolation process of NCDEs, i.e., our neural network-based interpolation vs. the existing explicit interpolation, and ii) exploit the generative characteristic of the decoder, i.e., extrapolation beyond the time domain of original data if needed. Therefore, our NCDE design can use both the interpolated and the extrapolated information for downstream machine learning tasks. In our experiments with 5 real-world datasets and 12 baselines, our extrapolation and interpolation-based NCDEs outperform existing baselines by non-trivial margins.
\end{abstract}

\ccsdesc[500]{Computing methodologies~Deep learning}
\ccsdesc[500]{Computing methodologies~Machine learning}
\ccsdesc[500]{Computing methodologies~Neural networks}
\ccsdesc[500]{Computing methodologies~Time Series}


\keywords{time-series data, neural controlled differential equations, extrapolation, interpolation}


\maketitle

\section{Introduction}
Deep learning for time-series data is popular for many web applications ~\cite{sezer2020financial,yan2018financial,yin2016forecasting,torres2017deep}, and many novel concepts have been proposed~\cite{busseti2012deep,fawaz2019deep,lim2021time}, ranging from recurrent neural networks (RNNs) to neural ordinary differential equations (NODEs~\cite{NIPS2018_7892}) and neural controlled differential equations (NCDEs~\cite{NEURIPS2020_4a5876b4}).

\begin{figure}
\centering
\includegraphics[width=0.8\columnwidth]{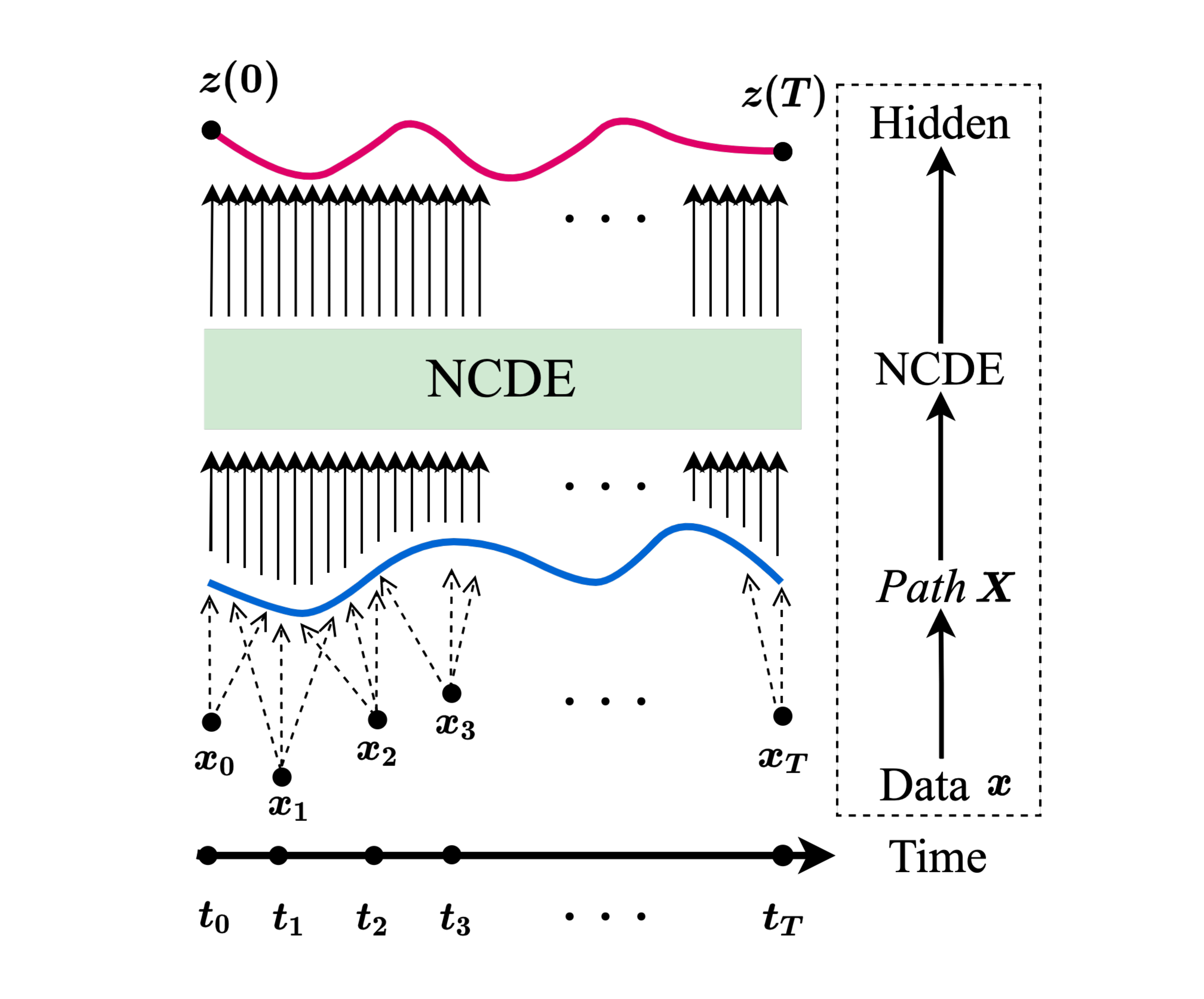}
\caption{The architecture of NCDE}
\label{fig:archi}
\end{figure}

\begin{figure}
\centering
\subfigure[Example of EXIT-NCDE, where the last integral time $\tau_{end}$ exceeds the original time domain of data from 0 to $T$]{\includegraphics[width=1\columnwidth]{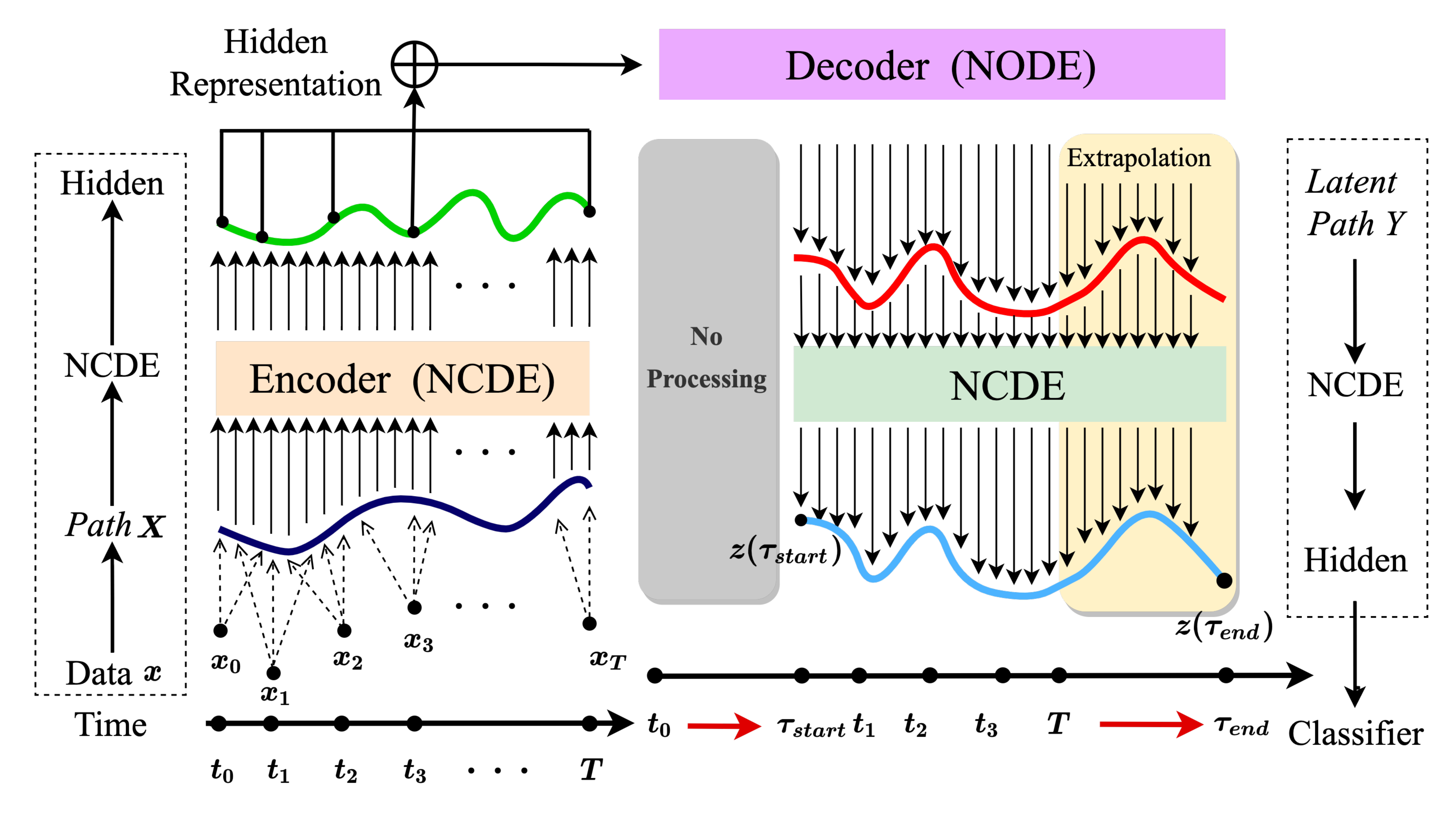}}
\subfigure[Example of EXIT-NCDE, where the last integral time $\tau_{end}$ is within the original time domain of data from 0 to $T$]{\includegraphics[width=1\columnwidth]{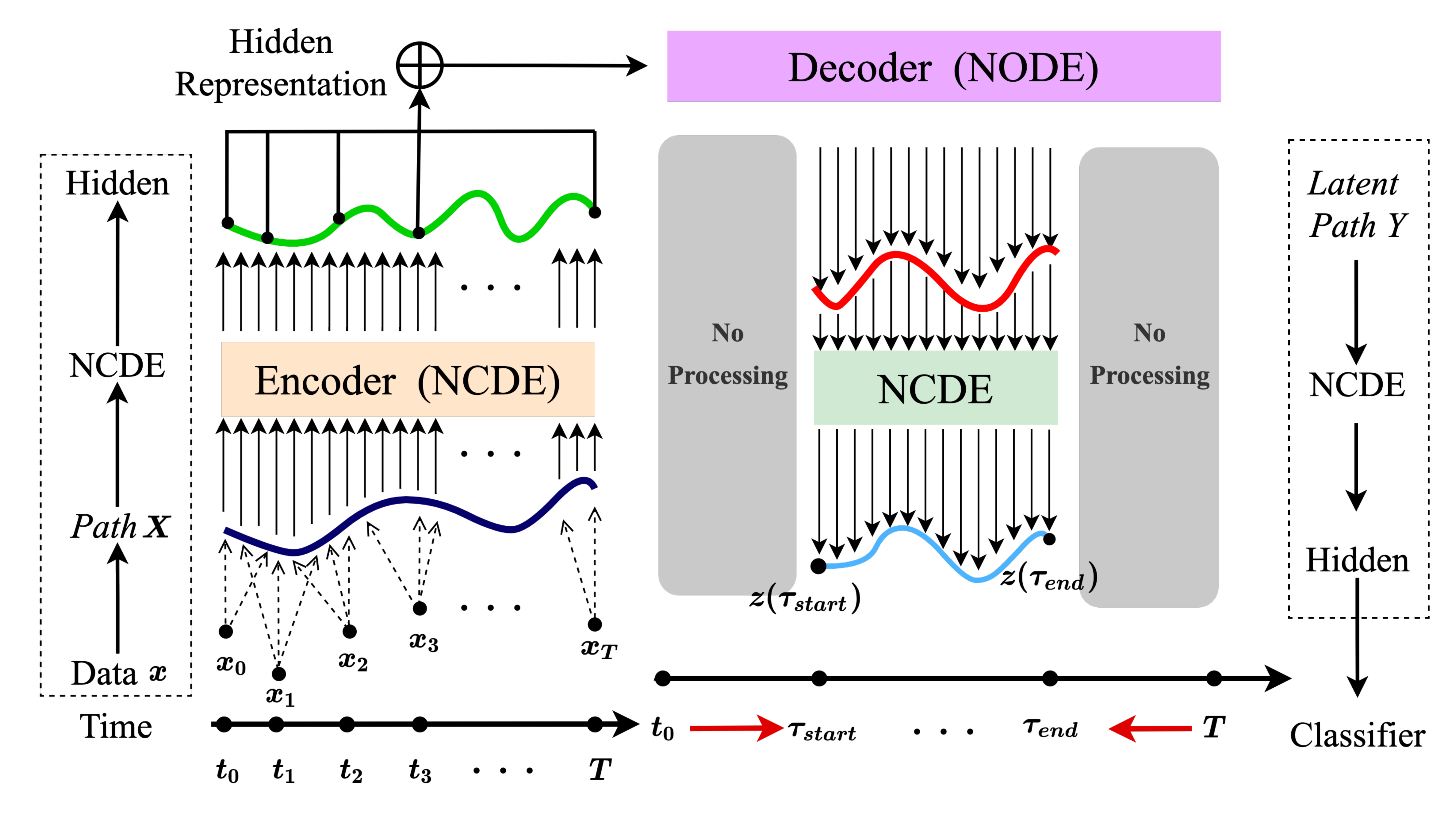}}
\caption{Examples of our proposed EXIT-NCDE. The key point in our model is that i) the latent path $Y$ is used rather than relying on the path $X$ created by an interpolation algorithm from data, and ii) the integral time duration $[\tau_{start}, \tau_{end}]$ is also trained. Given the condition of $0 \leq \tau_{start} < \tau_{end} < \infty$, there exist many possible cases, but these two cases the most frequently happen.}
\label{fig:archi2}
\end{figure}

RNNs have been typically used to process time-series data in the field of deep learning. Long short-term memory (LSTM~\cite{sepp1997long}) and gated recurrent unit (GRU~\cite{chung2014empirical}) are two representative RNN models, and they have resolved many technical issues of RNNs, ranging from the vanishing gradient problem to the complexity in processing time-series data. However, NODEs and NCDEs recently proposed breakthrough methods to process time-series data. Their advantages over RNNs can be summarized as follows: i) NODEs and NCDEs assume \emph{continuous} time and therefore, they are robust to irregular time-series data. RNNs are not suitable for processing time-series data with irregularity and/or missing information. ii) Much real-world time-series data can be well described by differential equations, e.g., the Black--Scholes differential equation describing the dynamics of a financial market~\cite{10.2307/1831029}. Therefore, modeling time-series data with NODEs and NCDEs are natural approaches. NODEs and NCDEs are formally written as follows:
\begin{enumerate}
    \item For NODEs,
    \begin{align}\label{eq:node}
    \bm{z}(T) = \bm{z}(0) + \int_{0}^{T} f(\bm{z}(t), t;\bm{\theta}_f) dt;
    \end{align}
    \item For NCDEs,
    \begin{align}\label{eq:ncde}
    \bm{z}(T) &= \bm{z}(0) + \int_{0}^{T} f(\bm{z}(t);\bm{\theta}_f) dX(t),\\
     &=\bm{z}(0) + \int_{0}^{T} f(\bm{z}(t);\bm{\theta}_f) \frac{dX(t)}{dt} dt,\label{eq:ncde2}
    \end{align}
    where $X(t)$ is a continuous path created by an interpolation algorithm
    from a raw discrete time-series sample by an interpolation algorithm (cf. Fig.~\ref{fig:archi}). We typically use the natural cubic spline~\cite{mckinley1998cubic} method to define the continuous path $X$ from a raw discrete time-series sample $\{(\bm{x}_i, t_i)\}_{i=1}^{N}$, where $\bm{x}_i$ means an observed values in the vector form --- we use boldface to denote vectors --- and $t_i \in [0,T]$ is its observation time. We note that $t_0 = 0$, $t_N = T$, and $t_i < t_{i+1}$.
    \item In both schemes, $\bm{z}(0)$ is an initial state vector, and $\bm{z}(t)$, where $t \in [0, T]$ is a state vector at time $t$. Therefore, they both describe how $\bm{z}(t)$ evolves over time $t \in [0, T]$.
\end{enumerate}

The theory of the controlled differential equation (CDE) had been developed to extend the stochastic differential equation and the It\^{o} calculus~\cite{cont2013functional} far beyond the semimartingale setting of $X$ --- in other words, Eq.~\eqref{eq:ncde} reduces to the stochastic differential equation if and only if $X$ meets the semimartingale requirement~\cite{protter1985approximations}. For instance, a prevalent example of the path $X$ is a Wiener process in the case of the stochastic differential equation. In CDEs, however, the path $X$ does not need to be such semimartingale or martingale processes. NCDEs are a technology to parameterize such CDEs and learn from data. In addition, Eq.~\eqref{eq:ncde2} continuously reads the values $\frac{dX(t)}{dt}$ and integrates them over time. In this regard, NCDEs are equivalent to continuous RNNs and show the state of the art accuracy in many time-series tasks and datasets.

In addition, how to create the continuous path $X$ from discrete time-series is one key part of NCDEs. As a matter of fact, NCDEs' model accuracy fluctuates depending on the interpolation methods ~\cite{morrill2021neural}. In this paper, we extend the interpolation-based NCDE model to an \underline{\textbf{EX}}trapolation and \underline{\textbf{I}}n\underline{\textbf{T}}erpolation-based model, called \emph{EXIT-NCDE}. Our method builds another latent path $Y$ from $X$ using an encoder-decoder architecture. The path $Y$ created by the decoder does not have limitation on its time domain whereas $X$ can be defined only in $[0,T]$. After that, we have the main NCDE that is defined on top of the latent path $Y$ and the integral time duration can be adjusted beyond $[0,T]$ as follows --- note that in Eqs.~\eqref{eq:node} and ~\eqref{eq:ncde} the integral time duration is always $[0,T]$:
\begin{align}
\bm{z}(\tau_{end}) &= \bm{z}(\tau_{start}) + \int_{\tau_{start}}^{\tau_{end}} g(\bm{z}(t);\bm{\theta}_g) dY(t),
\end{align}where the integral time duration $[\tau_{start}, \tau_{end}]$ is also trained. Therefore, our proposed method, EXIT, relies on both the interpolation and the extrapolation of time-series data (cf. Fig.~\ref{fig:archi2}). The benefits of learning $[\tau_{start}, \tau_{end}]$ can be summarized as follows:
\begin{enumerate}
\item An integral time duration of $[\tau_{start}, \tau_{end}]$, where $0 \leq \tau_{start} < \tau_{end} < \infty$, will be learned. However, the two most popular cases are as follows:
\begin{enumerate}
\item When much information is needed for a challenging downstream task, the decoder in EXIT can produce a longer path with an extended final integral time of $\tau_{end}$, i.e., $0 \leq \tau_{start} < T < \tau_{end}$.
\item When it is enough to consider only a subset of the latent path $Y$, the learned duration $[\tau_{start}, \tau_{end}]$ will be a subset of the original data time domain $[0, T]$, i.e., $0 < \tau_{start} < \tau_{end} < T$. In this way, we can prevent the \emph{overthinking} behavior of the main NCDE.
\end{enumerate}
\item We refer to Section~\ref{sec:mot} for the detailed motivations of learning the integral time duration.
\end{enumerate}

We conduct experiments with 5 benchmark time-series datasets for both classification and forecasting. We compare our method with various types of the state of the art methods, ranging from conventional RNNs to NODE and NCDE-based methods.


\section{Related Work}
We review time-series data processing. We first review RNN-based models and then NODE and NCDE-based models.

\paragraph{RNN-based Models} Vanilla RNNs have been widely used to model online streaming data such as natural language text, speech, and time series data ~\cite{zia2019long,zaremba2014recurrent,728168}. 




LSTMs are a model that improves the problem of vanilla RNNs, such as gradient extinction and explosion, and inability to capture sequential data among long-term dependencies. By solving the long-term dependency problem, LSTMs can process more macroscopic time data than RNNs. LSTMs include an internal memory cell state and 3 gates, i.e., a forget gate, an input gate, and an output gate, for storing long-term dependencies.


While LSTMs have the advantage of being able to capture long term dependencies, they have the disadvantage of having more parameters than vanilla RNNs. GRUs, which address the shortcomings of LSTMs, are a lightweight version of LSTMs and consist of two gates, i.e., a reset gate, and an update gate. In many cases, therefore, GRUs work as well as LSTMs.
 

\paragraph{NODE-based Models} NODEs use Eq.~\eqref{eq:node} to derive $\bm{z}(T)$ from $\bm{z}(0)$, where $f$ parameterized by $\bm{\theta}_f$ approximates $\dot{\bm{z}}(t) = \frac{d\bm{z}(t)}{dt}$. To solve the integral problem, we use ODE solvers and there exist various methods.

\begin{figure}
    \centering
    \includegraphics[width=1\columnwidth]{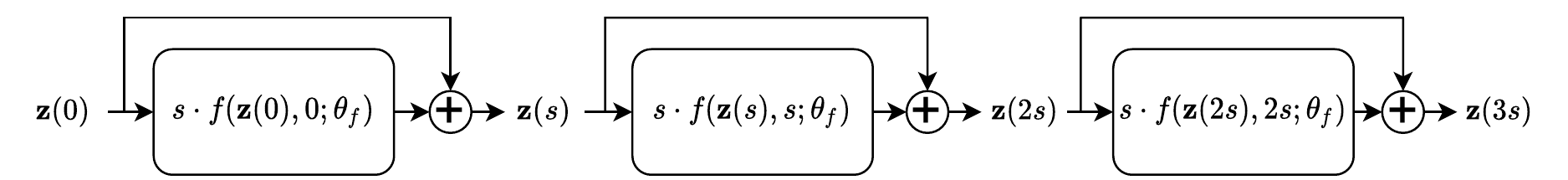}
    \caption{An illustration of the explicit Euler method to solve ODEs. Note that the entire workflow is analogue to ResNet ~\cite{he2016deep}.}
    \label{fig:euler}
\end{figure}

ODE solvers discretize the integral time domain $[0,T]$ in Eq.~\eqref{eq:node} into many small steps, and convert the integral into many steps of additions. For instance, the explicit Euler method can be written as follows in a step:
\begin{align}\label{eq:euler}
\bm{z}(t + s) = \bm{z}(t) + s \cdot f(\bm{z}(t), t;\bm{\theta}_f),
\end{align}where $s$, which is usually smaller than 1, is a configured step size of the Euler method. Note that this equation is identical to a residual connection when $s=1$ (cf. Fig.~\ref{fig:euler}).

Other ODE solvers use more complicated methods to update $\bm{z}(t + s)$ from $\bm{z}(t)$. For instance, the fourth-order Runge--Kutta (RK4) method uses the following method~\cite{butcher1976implementation}:
\begin{align}\label{eq:rk4}
\bm{z}(t + s) = \bm{z}(t) + \frac{s}{6}\Big(f_1 + 2f_2 + 2f_3 + f_4\Big),
\end{align}where $f_1 = f(\bm{z}(t), t;\bm{\theta}_f)$, $f_2 = f(\bm{z}(t) + \frac{s}{2}f_1, t+\frac{s}{2};\bm{\theta}_f)$, $f_3 = f(\bm{z}(t) + \frac{s}{2}f_2, t+\frac{s}{2};\bm{\theta}_f)$, and $f_4 = f(\bm{z}(t)+sf_3, t+s;\bm{\theta}_f)$.

The explicit Euler method is one of the most simplest ODE solvers. In addition to them, the Dormand--Prince (DOPRI) method is one of the most advanced solvers ~\cite{dormand1980family}. Whereas the explicit Euler method and RK4 use a fixed step-size, DOPRI uses an adaptive step-size. Therefore, we typically rely on DOPRI.

\begin{figure}
    \centering
    \includegraphics[width=0.8\columnwidth]{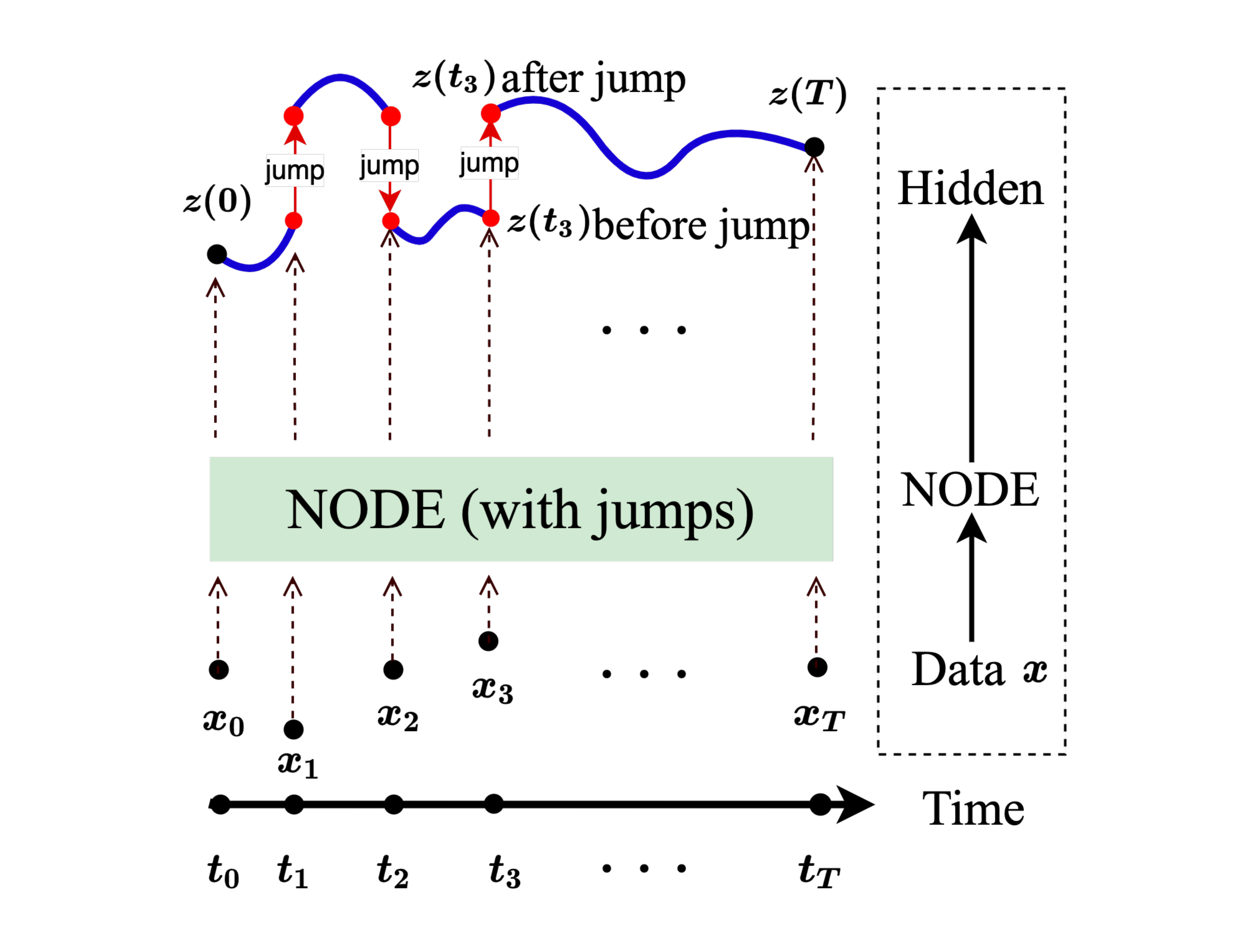}
    \caption{An example of how to process time-series with jump-based NODEs. Note that at each observation, $\bm{z}(t_i)$ jumps to another position considering the new input $\bm{x}_i$.}
    \label{fig:nodejump}
\end{figure}

In addition, NODEs have a breakthrough in its training method. Instead of the backpropagation, to train NODEs, we typically use the adjoint sensitivity method~\cite{NIPS2018_7892}. After letting $\bm{a}_{\bm{h}}(t) = \frac{d L}{d \bm{h}(t)}$ for a task-specific loss $L$, it calculates the gradient of loss w.r.t model parameters with another reverse-mode integral as follows:\begin{align}\label{eq:adj}
\nabla_{\bm{\theta}_f} L = \frac{d L}{d \bm{\theta}_f} = -\int_{T}^{0} \bm{a}_{\bm{h}}(t)^{\mathtt{T}} \frac{\partial f(\bm{h}(t), t;\bm{\theta}_f)}{\partial \bm{\theta}_f} dt.
\end{align}

$\nabla_{\bm{h}(0)} L$ can also be calculate in a similar way and we can propagate the gradient backward to the layers earlier than the ODE if any. It is worth of mentioning that the space complexity of the adjoint sensitivity method is $\mathcal{O}(1)$ whereas using the backpropagation to train NODEs has a space complexity proportional to the number of DOPRI steps. Their time complexities are similar or the adjoint sensitivity method is slightly more efficient than that of the backpropagation. Therefore, we can train NODEs efficiently. Recently, moreover, two more advanced training methods for NODEs have been proposed: the adaptive checkpoint adjoint (ACA)~\cite{zhuang2020adaptive} method and the memory-efficient ALF integrator (MALI)~\cite{zhuang2021mali}. These methods greatly stabilizes the training process of NODEs.

Although NODEs can handle time-series data directly, we usually use more advanced architectures such as Latent-ODE ~\cite{NIPS2019_8773}, GRU-ODE ~\cite{debrouwer2019gruodebayes,jordan2021gated}, and so on. A time-series model in which the latent state follows a NODE is known as a latent-ODE. GRU-ODE is a model that continuously evolves the state by modeling GRUs with NODEs. Fig.~\ref{fig:nodejump} depicts a GRU-ODE with jumps.

\paragraph{NCDE-based Models}
NCDEs are considered more advanced than NODEs since NODEs are theoretically a special case of NCDEs --- e.g., Eq.~\eqref{eq:ncde} reduces to Eq.~\eqref{eq:node} when $\frac{dX(t)}{dt} = 1$. In NCDEs, therefore, the path $X$ is crucial. Because of the existence of the path $X$, Kidger et al. prove that NCDEs can learn what NODEs cannot (See Theorem C.1 in~\cite{NEURIPS2020_4a5876b4}). When the path $X$ is a Wiener process, Eq.~\eqref{eq:ncde} is called \emph{stochastic differential equations} (SDEs~\cite{tzen2019neural}). Therefore, CDEs are a broader concept that subsumes SDEs. As a matter of fact, the theory of CDE had been developed to eradicate the restriction that the path $X$ should a semimartingale process in SDEs~\cite{lyons2002system}.

Despite the fundamental difference between NODEs and NCDEs, the final form of Eq.~\eqref{eq:ncde} is derived to Eq.~\eqref{eq:ncde2} which can be solved by existing ODE solvers. In Eq.~\eqref{eq:ncde2}, $\dot{\bm{z}}(t)$ is modeled by $f(\bm{z}(t);\bm{\theta}_f) \frac{dX(t)}{dt}$, which is not the case in NODEs. However, existing ODE solvers can somehow solve the problem once we can successfully model $\dot{\bm{z}}(t)$, which is a technical circumvent to deal with NCDEs with existing technology. We can also apply the ACA or the MALI algorithm to train NCDEs for the same reason. In addition, the calculation of $f(\bm{z}(t);\bm{\theta}_f) \frac{dX(t)}{dt}$ is a matrix-vector multiplication which can be done quickly. In general, NCDEs do not significantly increase the computational complexity in comparison with NODEs but they can theoretically learn better than NODEs (since NODEs are a special case of NCDEs).

Surprisingly, NCDE-based models outperform NODE-based models for time-series tasks without advanced architectures such as an encoder-decoder style model. Recently, however, one method to enhance NCDEs with an attention mechanism, called ANCDE, has been proposed in~\cite{jhin2021ancde}.

\section{Proposed Method}
We describe our design to enhance NCDEs with both interpolation and extrapolation. The overall architecture is first described, followed by detailed descriptions.

\begin{table}[]
\small
\setlength{\tabcolsep}{2pt}
\caption{All possible cases of $\tau_{start}$}\label{tbl:tau1}
\begin{tabular}{cc}
\hline
Learned Value & Meaning \\ \hline
$\tau_{start}=0$ & \begin{tabular}[c]{@{}c@{}}Earlier values are important for a downstream task.\end{tabular} \\ 
$\tau_{start} > 0$ & \begin{tabular}[c]{@{}c@{}}Earlier values can be ignored for a downstream task.\end{tabular} \\ \hline
\end{tabular}
\end{table}

\begin{table}[]
\small
\setlength{\tabcolsep}{2pt}
\caption{All possible cases of $\tau_{end}$}\label{tbl:tau2}
\begin{tabular}{cc}
\hline
Learned Value & Meaning \\ \hline
$\tau_{end} < T$ & \begin{tabular}[c]{@{}c@{}}Later values can be ignored for a downstream task.\end{tabular} \\ 
$\tau_{end} = T$ & \begin{tabular}[c]{@{}c@{}}Later values are important for a downstream task.\end{tabular} \\ 
$\tau_{end} > T$ & \begin{tabular}[c]{@{}c@{}}More information is needed for a downstream task.\end{tabular} \\ \hline
\end{tabular}
\end{table}

\subsection{Motivations}\label{sec:mot}
Given a time-series sample $\{(\bm{x}_i, t_i)\}_{i=1}^{N}$, the original NCDE design builds an interpolated path $X$ in the pre-determined time domain $[0,T]$. However, it is unclear whether this scheme is the optimal method or not. To this end, we let our encoder-decoder networks create another latent path $Y$ (from $X$) whose time domain is unlimited, i.e., $[0,\infty]$. Then, we also train the integral time duration, denoted $[\tau_{start}, \tau_{end}]$ in our paper, where $0 \leq \tau_{start} < \tau_{end} < \infty$. All possible cases are summarized in Tables~\ref{tbl:tau1} and~\ref{tbl:tau2}. We observe in our experiments that i) sometimes $\tau_{end}$ gets larger beyond $T$ and other times smaller below $T$, depending on datasets, and ii) $\tau_{start}$ is mostly trained larger than 0 and rarely stays on 0.

The motivation of learning $[\tau_{start}, \tau_{end}]$ is that it is sometimes need to feed much information for the main NCDE model (e.g., the NCDE model highlighted in a green box in Fig.~\ref{fig:archi2} (a)) to accomplish a challenging downstream task. Since our encoder-decoder architecture produces another latent path $Y$ for the time domain $[0,\infty]$, we can feed as much information as needed into the main NCDE model. Moreover, the encoder-decoder architecture is trained to produce such a reliable long path.

Another motivation is that deep neural networks are frequently blamed for \emph{overthinking}, resulting in overfitting and longer inference time ~\cite{kaya2019shallow}. In order to prevent such a problem, we need to make deep neural networks \emph{shallow} in terms of their layers, which corresponds to decreasing the integral time duration $[\tau_{start}, \tau_{end}]$ smaller. For instance, the number of steps in Fig.~\ref{fig:euler} will decrease by making the integral time duration small.

In addition, it is needed to train $\tau_{start}, \tau_{end}$ since we do not know what are the optimal settings for them for a given dataset. In our method, therefore, we learn $\tau_{start}, \tau_{end}$ rather than fixing to a predetermined value.

\subsection{Overall Workflow}\label{sec:flow}
The overall workflow in our model is as follows --- its schematic diagram is in Fig.~\ref{fig:flow}:
\begin{enumerate}
    \item Given a discrete time-series sample $\{(\bm{x}_i, t_i)\}_{i=1}^{N}$, there is an interpolation algorithm, i.e., the natural cubic spline algorithm in our case, which produces a continuous path $X$. We note that $X(t_i) = \bm{x}_i$ for each observed time-point $t_i$. For other non-observed time-points, the interpolation algorithm fills out appropriate values.
    \item Our NCDE-based encoder reads the path $X$ to produce its hidden representations $\{\bm{e}(t_i)\}_{i=1}^{N}$, which will be concatenated into a large hidden vector representing the entire path $X$.
    \item From the concatenated hidden vector, there is a NODE-based decoder which produces another latent path $Y$.
    \item The main NCDE reads the latent path $Y$ and produces the last hidden vector $\bm{z}(\tau_{end})$. There is an output layer (omitted in Fig.~\ref{fig:flow}) to process the last hidden vector and make inference.
\end{enumerate}

\subsection{Encoder-Decoder to Build Latent Path $Y$}
We propose to evolve $\bm{z}$ beyond the final observation time $T$, if needed, owing to the NODE-based decoder's extrapolation capability. Therefore, our proposed NCDE framework relies on both the interpolation (when $\tau_{end} \leq T$) and the extrapolation (when $\tau_{end} > T$). Moreover, the entire integral time duration $[\tau_{start}, \tau_{end}]$ is trained rather than being fixed. Therefore, our method can be written as follows:
\begin{align}
    \bm{z}(\tau_{end}) &= \bm{z}(\tau_{start}) + \int_{\tau_{start}}^{\tau_{end}} g(\bm{z}(t);\bm{\theta}_g) \frac{dY(t)}{dt} dt,\label{eq:exttop}\\
    Y(\tau_{end}) &= Y(\tau_{start}) + \int_{\tau_{start}}^{\tau_{end}} f(Y(t), t;\bm{\theta}_f) dt,\label{eq:mid}\\
    \bm{e}(T) &= \bm{e}(0) + \int_{0}^{T} k(\bm{e}(t);\bm{\theta}_k) \frac{dX(t)}{dt} dt,\label{eq:extbottom}
\end{align} where $Y(\tau_{start}) = \phi_{Y}(\oplus_i\bm{e}(t_i)
;\bm{\theta}_{\phi_{Y}})$, $\bm{z}(\tau_{start}) = \phi_{\bm{z}}(X(\tau_{start});\bm{\theta}_{\phi_{\bm{z}}})$, $\phi_{Y}$ and $\phi_{\bm{z}}$ are two fully-connected (FC) layer-based mapping functions, and $\oplus$ means the concatenation operator. The NCDE of $\bm{e}(t)$ can be considered as an encoder to produce a set of hidden representations $\{\bm{e}(t_i)\}_{i=1}^N$ given a time-series sample $\{(\bm{x}_i, t_i)\}_{i=1}^{N}$. Then, there is a NODE-based decoder, denoted $Y(t)$, which produces another latent path in $[0, \infty]$. $\bm{z}(t)$ can be written as follows after combining Eqs.~\eqref{eq:exttop} and~\eqref{eq:mid}:
\begin{align}
    \bm{z}(\tau_{end}) &= \bm{z}(\tau_{start}) + \int_{\tau_{start}}^{\tau_{end}} g(\bm{z}(t);\bm{\theta}_g) f(Y(t), t;\bm{\theta}_f) dt.\label{eq:exttop2}
\end{align}

We note that Eq.~\ref{eq:exttop2} is equivalent to Eq.~\eqref{eq:exttop}, but for our convenience, we implement Eq.~\ref{eq:exttop2}. The ODE and CDE functions $g,f,k$ for time-series classification are summarizes in Tables~\ref{tbl:f1} and~\ref{tbl:f2}. Those functions for time-series forecasting are summarized in Appendix.

\begin{table}[t]
\scriptsize
\setlength{\tabcolsep}{4pt}
\caption{The best architecture of the CDE function $k$ for Time Series Classification. \texttt{FC}, $\rho$, and $\xi$ stands for the fully-connected layer, the rectified linear unit (ReLU), and the hyperbolic tangent (tanh), respectively.}\label{tbl:f1}
\begin{tabular}{cccccccc}
\hline
\multirow{2}{*}{Design} & \multirow{2}{*}{Layer} & \multicolumn{2}{c}{PhysioNet Sepsis} & \multicolumn{2}{c}{Character Trajectories} & \multicolumn{2}{c}{Speech Commands} \\ \cline{3-8} 
                        &                        & Input             & Output           & Input                & Output              & Input            & Output           \\ \hline
\texttt{FC}                      & 1                      & $1024 \times $49         & $1024 \times $69        & $32 \times $40              & $32 \times $90             & $1024 \times $60        & $1024 \times $100       \\
$\rho$(\texttt{FC})                      & 2                      & $1024 \times $69         & $1024 \times $69        & $32 \times $90              & $32 \times $90             & $1024 \times $100       & $1024 \times $100       \\
$\rho$(\texttt{FC})                      & 3                      & $1024 \times $69         & $1024 \times $69        & $32 \times $90              & $32 \times $90             & $1024 \times $100       & $1024 \times $100       \\
$\rho$(\texttt{FC})                      & 4                      & --                & --               & --                   & --                  & $1024 \times $100       & $1024 \times $100       \\
$\xi$(\texttt{FC})                      & 5                      & $1024 \times $69         & $1024 \times $3381        & $32 \times $90              & $32 \times $160             & $1024 \times $100       & $1024 \times $1260        \\ \hline
\end{tabular}
\end{table}
\begin{table}[t]
\scriptsize
\setlength{\tabcolsep}{4pt}
\caption{The best architecture of the CDE function $g$ and the ODE function $f$  for Time Series Classification}\label{tbl:f2}
\begin{tabular}{cccccccc}
\hline
\multirow{2}{*}{Design} & \multirow{2}{*}{Layer} & \multicolumn{2}{c}{PhysioNet Sepsis} & \multicolumn{2}{c}{Character Trajectories} & \multicolumn{2}{c}{Speech Commands} \\ \cline{3-8} 
                        &                        & Input             & Output           & Input                & Output              & Input            & Output           \\ \hline
\texttt{FC}                      & 1                      & $1024 \times $49         & $1024 \times $69        & $32 \times $40              & $32 \times $90             & $1024 \times $60        & $1024 \times $100       \\
$\rho$(\texttt{FC})                      & 2                      & $1024 \times $69         & $1024 \times $69        & $32 \times $90              & $32 \times $90             & $1024 \times $100       & $1024 \times $100       \\
$\rho$(\texttt{FC})                      & 3                      & $1024 \times $69         & $1024 \times $69        & $32 \times $90              & $32 \times $90             & $1024 \times $100       & $1024 \times $100       \\
$\rho$(\texttt{FC})                      & 4                      & --                & --               & --                   & --                  & $1024 \times $100       & $1024 \times $100       \\
$\xi$(\texttt{FC})                      & 5                      & $1024 \times $69         & $1024 \times $49        & $32 \times $90              & $32 \times $40             & $1024 \times $100       & $1024 \times $60        \\ \hline
\end{tabular}
\end{table}


\paragraph{\textbf{How to train $\tau_{start},\tau_{end}$}} At the beginning of training process, we initialize $\tau_{start}$ to 0 and $\tau_{end}$ to $T$. We also impose the relationship $0 \leq \tau_{start} < \tau_{end} < \infty$ in our source codes by using appropriate APIs, such as \texttt{torch.clamp} and so on. To train $\tau_{end}$, we perform $\tau_{end} = \tau_{end} - \lambda_{\tau} \frac{d L}{d \tau_{end}}$, where $\lambda_{\tau}$ is a learning rate and $L$ is a task loss, with the following gradients:
\begin{align}\begin{split}\label{eq:tau1}
\frac{d L}{d \tau_{end}} &= \frac{d L}{d \bm{z}(\tau_{end})}\frac{d \bm{z}(\tau_{end})}{d \tau_{end}}\\
&= \frac{d L}{d \bm{z}(\tau_{end})} \Big(g(\bm{z}(\tau_{end});\bm{\theta}_g) f(Y(\tau_{end}), \tau_{end};\bm{\theta}_f)\Big)\\
&=\bm{a}_{\bm{z}}(\tau_{end}) \Big(g(\bm{z}(\tau_{end});\bm{\theta}_g) f(Y(\tau_{end}), \tau_{end};\bm{\theta}_f)\Big),
\end{split}\end{align}where $\bm{a}_{\bm{z}}(\tau_{end})$ is an adjoint state at time $\tau_{end}$ which can be easily calculated with existing ODE solvers as in Eq.~\eqref{eq:adj}.

To train $\tau_{start}$, we use a similar method with the following gradient definition:
\begin{align}\begin{split}\label{eq:tau2}
\frac{d L}{d \tau_{start}} &= \frac{d L}{d \bm{z}(\tau_{start})} \Big(g(\bm{z}(\tau_{start});\bm{\theta}_g) f(Y(\tau_{start}), \tau_{start};\bm{\theta}_f)\Big)\\
&= \bm{a}_{\bm{z}}(\tau_{start}) \Big(g(\bm{z}(\tau_{start});\bm{\theta}_g) f(Y(\tau_{start}), \tau_{start};\bm{\theta}_f)\Big).
\end{split}\end{align}

We note that in most cases of our experiments, $\tau_{start}$ is trained to be larger than 0. $\tau_{end}$ sometimes gets larger beyond the physical time domain $[0,T]$ since the path $Y$ is in a latent space and there is no limitation on its latent time domain. However, we observe from our experiments that in other cases, $\tau_{end}$ can get smaller than $T$. All these results are case by case. Since $[\tau_{start}, \tau_{end}]$ is trained in our case, it will converge to a reasonable time duration, where the task loss is optimized.

\paragraph{\textbf{Rationale behind learning $[\tau_{start}, \tau_{end}]$}} The relationship between $\tau_{start}$ and $\tau_{end}$ is $0 \leq \tau_{start} < \tau_{end} < \infty$, and they can converge to any values satisfying the relationship. Among many possible cases, we found that in our experiments, the two cases exemplified in Fig.~\ref{fig:archi2} happen the most frequently.

By increasing $\tau_{end}$ beyond $T$, we can feed more information to our main NCDE model to accomplish a downstream task. This happens when the task is challenging, where the challenging workload is shared by the two NCDE and one NODE models, denoted $\bm{e}(t)$, $Y(t)$, and $\bm{z}(t)$, respectively.

By decreasing $\tau$ below $T$, on the other hand, we can prevent the problems of \emph{overthinking} and \emph{overfeeding}, where neural networks process too much information with too much processing capabilities, frequently resulting in overfitting.

\begin{figure}
    \centering
    \includegraphics[width=1\columnwidth]{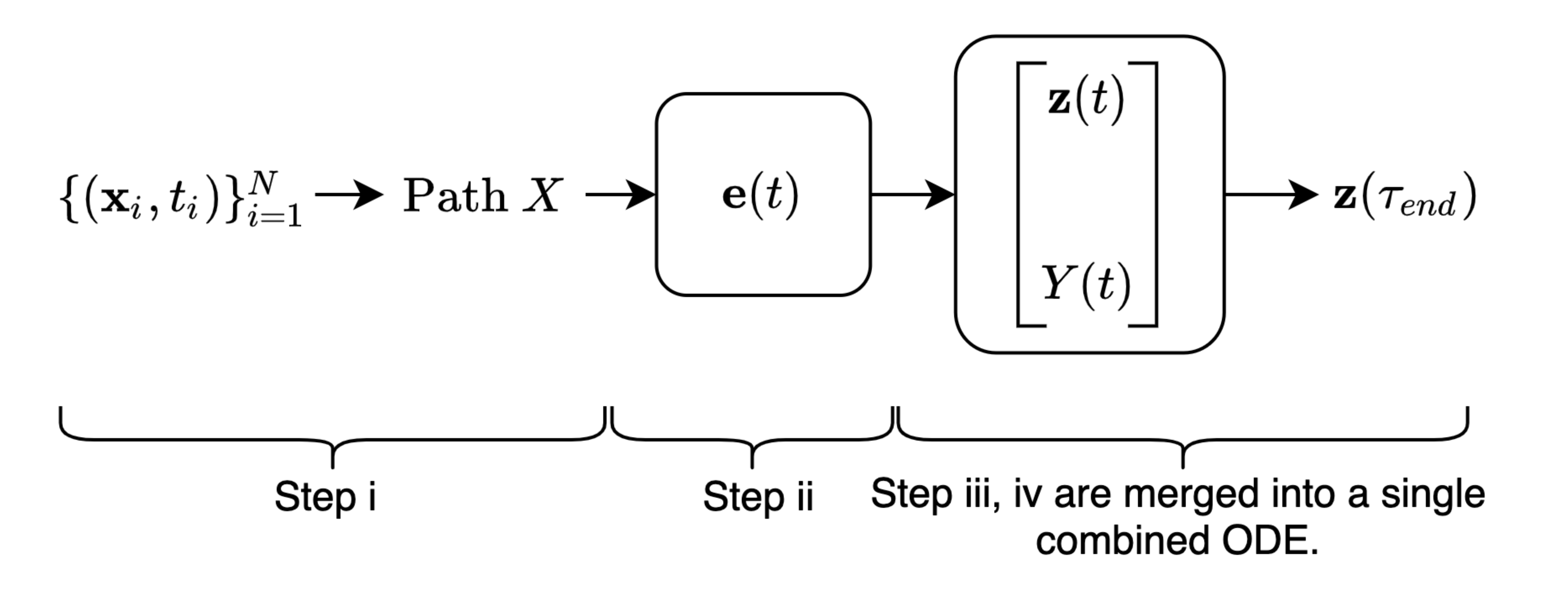}
    \caption{The block diagram of the processing sequence in our proposed method. See Section~\ref{sec:flow} for the detailed descriptions of the steps.}
    \label{fig:flow}
\end{figure}

\paragraph{\textbf{How to implement}} Fig.~\ref{fig:flow} shows how we implement our proposed method. To implement in an efficient way, we define the following combined ODE:
\begin{align}
\frac{d}{dt}{\begin{bmatrix}
  \bm{z}(t) \\
  Y(t) \\
  \end{bmatrix}\!} = {\begin{bmatrix}
  g(\bm{z}(t);\bm{\theta}_g) f(Y(t), t;\bm{\theta}_f) \\
  f(Y(t), t;\bm{\theta}_f)\\
  \end{bmatrix}\!},
\end{align}where the two different evolving processes of $\bm{z}(t)$ and $Y(t)$ are merged into a single ODE. Since both $\frac{d \bm{z}(t)}{dt}$ and $\frac{d Y(t)}{dt}$ refer to the ODE function $f$, we do not need to separate them but implement in a combined manner as above.

In addition, it is also important how to define the initial values of $\bm{z}(0)$ and $Y(0)$, and we use the following method:
\begin{align}
{\begin{bmatrix}
  \bm{z}(0) \\
  Y(0) \\
  \end{bmatrix}\!} = {\begin{bmatrix}
  \phi_{\bm{z}}(X(0);\bm{\theta}_{\phi_{\bm{z}}}) \\
  \phi_{Y}(\oplus_i\bm{e}(t_i) ;\bm{\theta}_{\phi_{Y}}) \\
  \end{bmatrix}\!}, 
\end{align}where $\phi_{\bm{z}}$ and $\phi_{Y}$ are two FC-based mapping functions.

\subsection{Training Method}
\paragraph{\textbf{Training algorithm}} We use the same training algorithm and the same output layer used by the original NCDE design~\cite{NEURIPS2020_4a5876b4}. Given the last hidden vector $\bm{z}(\tau_{end})$, there is a fully connected layer, followed by a softmax activation, for time-series classification. In the case of time-series forecasting, there is only one fully connected layer without any activations. We denote the trainable parameters of the output layer as $\bm{\theta}_{output}$. We use the cross entropy loss for classification and the mean squared error (MSE) loss for forecasting. We also use the kinetic regularization~\cite{finlay2020train} with a coefficient $c_{kr}$ to regularize $\bm{\theta}_f$, $\bm{\theta}_g$, $\bm{\theta}_k$, and the weight decay with a coefficient $c_{wd}$ to regularize $\bm{\theta}_{\phi_{\bm{z}}}$, $\bm{\theta}_{\phi_{\bm{Y}}}$, and $\bm{\theta}_{output}$. The training algorithm is in Alg.~\eqref{alg:train}.

\paragraph{\textbf{Well-posedness}} The well-posedness\footnote{A well-posed problem means i) its solution uniquely exists, and ii) its solution continuously changes as input data changes.} of NODEs/NCDEs was already proved in \cite[Theorem 1.3]{lyons2004differential} under the mild condition of the Lipschitz continuity. Almost all activations, such as ReLU, ELU, Leaky ReLU, SoftPlus, Tanh, Sigmoid, ArcTan, and Softsign, have a Lipschitz constant of 1. Other common neural network layers, such as dropout, batch normalization and other pooling methods, have explicit Lipschitz constant values. Therefore, the Lipschitz continuity of $g, f, k$ can be fulfilled in our case. This makes our training problem for NCDEs well-posed. As a result, our training algorithm solves a well-posed problem so its training process is stable in practice.

\begin{algorithm}[t]
\small
\SetAlgoLined
\caption{How to train EXIT-NCDE}\label{alg:train}
\KwIn{Training data $D_{train}$, Validating data $D_{val}$, Maximum iteration numbers $max\_iter$}
Initialize $\bm{\theta}_f$, $\bm{\theta}_g$, $\bm{\theta}_k$, $\bm{\theta}_{\phi_{\bm{z}}}$, $\bm{\theta}_{\phi_{\bm{Y}}}$, and $\bm{\theta}_{output}$;

$i \gets 0$;

\While {$i < max\_iter$}{
    Train $\bm{\theta}_f$, $\bm{\theta}_g$, $\bm{\theta}_k$, $\bm{\theta}_{\phi_{\bm{z}}}$, $\bm{\theta}_{\phi_{\bm{Y}}}$, and $\bm{\theta}_{output}$ using a task-oriented loss and a kinetic regularization with a learning rate of $\lambda$\;\label{alg:train3}
    
    Train $\tau_{start}, \tau_{end}$ with the gradients in Eqs.~\eqref{eq:tau1} and~\eqref{eq:tau2}, and a learning rate $\lambda_{tau}$\;
    
    Validate and update the best parameters, $\bm{\theta}^*_f$, $\bm{\theta}^*_g$, $\bm{\theta}^*_k$, $\bm{\theta}^*_{\phi_{\bm{z}}}$, $\bm{\theta}^*_{\phi_{\bm{Y}}}$, and $\bm{\theta}^*_{output}$, with $D_{val}$\;
    
    $i \gets i + 1$;
}
\Return $\bm{\theta}^*_f$, $\bm{\theta}^*_g$, $\bm{\theta}^*_k$, $\bm{\theta}^*_{\phi_{\bm{z}}}$, $\bm{\theta}^*_{\phi_{\bm{Y}}}$, and $\bm{\theta}^*_{output}$;
\end{algorithm}

\section{Experiments}
We conduct classification and forecasting experiments with 5 real-world datasets and 12 baselines. We run five times with five different seeds and report their statistical performance.

\paragraph{Baselines}
For time-series classification and forecasting experiments, we compare our method with the following baselines: RNN, LSTM, and GRU are all recurrent neural network-based models that can process sequential data. LSTM is designed to learn long-term dependencies by overcoming the shortcomings of RNNs, and GRU uses a gating mechanism to control the flow of information. GRU-ODE is a successor to NODE, which continuously models GRU as NODE. GRU-$\Delta t$ is a GRU model that additionally receives the time difference information between observations. GRU-D~\cite{che2018recurrent} is a modified version of GRU-$\Delta t$ with a learnable exponential decay between observations. ODE-RNN is an extension of GRU-$\Delta t$ based on NODEs. The combination of NODEs and GRUs is called a \emph{jump}. Latent-ODE is a good model for time-series when latent states can be described by ODEs. In this study, the recognition network of the existing Latent-ODE model is denoted as ODE-RNN. In this paper, Augmented-ODE~\cite{NIPS2019_8577} increased the ODE state size of Latent-ODE. ACE-NODE is one of the state of the art attention-based NODE model with a dual co-evolving NODEs. ANCDE is an attention-based NCDE model.

\paragraph{Hyperparameters}

For baselines, we refer to Appendix for their hyperparameter configurations --- we conduct another search on our own based on their recommended configurations. 
For our model, we consider the following hyperparameter configurations: the number of layers in the ODE/CDE functions is $\{2,3,4,5\}$, the dimensionality of hidden vector $\bm{h}(t)$ is $\{20,30,40,50,80,100\}$.
we use a learning rate $\lambda$ of $\{\num{1.0e-2}, \num{1.0e-4}, \num{5.0e-3}, \num{5.0e-4}\}$, and a learning rate $\lambda_{\tau}$ of $\{\num{1.0e-4}, \num{1.0e-3}, \num{1.0e-2},\num{1.0e-1}, \num{1.0}\}$

The best architectures for time-series classification are in Tables~\ref{tbl:f1} and~\ref{tbl:f2}. Other best hyperparameter information is in Appendix.

\subsection{Time Series Classification}
We introduce our experimental results for time-series classification with the following three datasets. We use the accuracy for balanced classification datasets and AUROC for imbalanced datasets.

\paragraph{PhysioNet Sepsis}
The PhysioNet 2019 challenge to predict sepsis~\cite{9005736,article} is one of the most popular irregular time-series classification experiments. The status of patients in the ICU --- both static and time-dependent features --- are recorded in this dataset, and only 34 time-dependent features are used for time-series classification. We predict the onset of sepsis through this classification. The dataset consists of 40,355 cases with variable time-series lengths, with approximately 90\% missing observations. Because of this irregularity in the data, we perform two types of time-series classification: i) time-series classification with observation intensity (OI) and ii) time-series classification excluding observation intensity (no OI). In the former case, the intensity, which is an index for each time-series observation, can suggest the degree of serious illness. Since the data itself is unbalanced, we use AUROC as a metric.

\paragraph{Speech Commands}
The Speech Commands dataset is one second long audio data recorded with spoken words (e.g., `left', `right', `cat', `dog') and background noise~\cite{DBLP:journals/corr/abs-1804-03209}. It has balanced all 35 labels with words `yes', `no', `up', `down', `left', `right', `on', `off', `stop', etc. It consists of 34,975 samples, each of which has a time-series length of 161 and has an input size of 20 dimensions.

\paragraph{Character Trajectories}
This dataset is used for classification and is one of the UEA time-series classification archive datasets~\cite{bagnall2018uea}. The x-axis, y-axis, and pen tip force values of the Latin alphabet were obtained using a tablet with a sample frequency of 200 Hz and the x-axis, y-axis, and pen tip force values of the Latin alphabet. It comprises of 2,858 samples, each with a time-series length of 182 and a three-dimensional input size.
The dataset contains in total 20 alphabetic classes (`a', `b', `c', `d', `e', `g', `h', `l', `m', `n', `o', `p', `q', `r', `s', `u', `v', `w', `y', `z') are used and the remaining 6 characters are excluded.

\paragraph{Experimental results}

\begin{table}[t]
\footnotesize
\setlength{\tabcolsep}{4pt}
\centering
\caption{AUROC (mean ± std, computed across five runs) on PhysioNet Sepsis. Memory is in megabytes.}\label{tbl:PhysioNet}
\begin{tabular}{cccccc} \hline
\multicolumn{1}{c}{\multirow{2}{*}{Model}} & \multicolumn{2}{c}{Test AUROC} & & \multicolumn{2}{c}{Memory Usage (MB)} \\ \cline{2-3}\cline{5-6}
\multicolumn{1}{c}{}  & OI                  & No OI                     && OI  & No OI  \\ \hline
GRU-$\Delta t$  & 0.878 ± 0.006             & 0.840 ± 0.007             && 837 & 826    \\
GRU-D           & 0.871 ± 0.022             & \textbf{0.850 ± 0.013}  && 889 & 878    \\
GRU-ODE         & 0.852 ± 0.010             & 0.771 ± 0.024             && 454 & 273    \\
ODE-RNN         & 0.874 ± 0.016             & 0.833 ± 0.020             && 696 & 686    \\
Latent-ODE      & 0.787 ± 0.011             & 0.495 ± 0.002             && 133 & 126    \\
Augmented-ODE   & 0.832 ± 0.015             & 0.497 ± 0.010             && 998 & 283    \\
ACE-NODE        & 0.804 ± 0.010             & 0.514 ± 0.003             && 194 & 218    \\
NCDE            & 0.880 ± 0.006             & 0.776 ± 0.009             && 244 & 122    \\
ANCDE           & 0.900 ± 0.002             & 0.823 ± 0.003             && 285 & 129    \\ \hline
\textbf{EXIT}   & \textbf{0.913 ± 0.002}    & 0.836 ± 0.003             && 257 & 127    \\ \hline
\end{tabular}
\end{table}

\begin{table}[t]
\begin{minipage}{.48\linewidth}
\setlength{\tabcolsep}{1pt}
\scriptsize
\centering
\caption{Accuracy on Speech Commands}\label{tbl:speech}
\begin{tabular}{ccc}
\hline
Model & Test Accuracy &Memory \\ 
\hline
RNN             &  0.197 ± 0.006             &1,905     \\
LSTM            &  0.684 ± 0.034             &4,080     \\
GRU             &  0.747 ± 0.050             &4,609     \\
GRU-$\Delta t$  &  0.433 ± 0.339             &1,612     \\
GRU-D           &  0.324 ± 0.348             &1,717     \\
GRU-ODE         &  0.479 ± 0.029             &171.3     \\
ODE-RNN         &  0.659 ± 0.356             &1,472     \\
Latent-ODE      &  0.920 ± 0.006             &2,668     \\
Augmented-ODE   &  0.913 ± 0.008             &2,626     \\
ACE-NODE        &  0.911 ± 0.003             &3,046     \\
NCDE            &  0.898 ± 0.025             &174.9     \\
ANCDE           &  0.807 ± 0.075             &179.8     \\
\hline  
\textbf{EXIT}   & \textbf{0.930 ± 0.003}     &178.5     \\
\hline
\end{tabular}
\end{minipage}\hfill
\begin{minipage}{.48\linewidth}
\setlength{\tabcolsep}{1pt}
\scriptsize
\centering
\caption{Accuracy on Regular Character Trajectories}\label{tbl:charactertrajectory1}
\begin{tabular}{ccc}
\hline
Model           & Test Accuracy           & Memory \\ \hline
RNN             & 0.211 ± 0.038           & 52.2    \\
LSTM            & 0.791 ± 0.113           & 48.6    \\
GRU             & 0.844 ± 0.079           & 54.8   \\
GRU-$\Delta t$  & 0.834 ± 0.132           & 16.5    \\
GRU-D           & 0.896 ± 0.050           & 17.8    \\
GRU-ODE         & 0.778 ± 0.091           & 1.51    \\
ODE-RNN         & 0.427 ± 0.078           & 15.5    \\
Latent-ODE      & 0.954 ± 0.003           & 181     \\
Augmented-ODE   & 0.970 ± 0.012           & 186     \\
ACE-NODE        & 0.981 ± 0.001           & 113     \\
NCDE            & 0.974 ± 0.004           & 1.38    \\
ANCDE           & 0.991 ± 0.002           & 2.02    \\
\hline
\textbf{EXIT} & \textbf{0.993 ± 0.001}  & 3.79\\
\hline
\end{tabular}
\end{minipage}
\end{table}

\begin{table}[t]
\footnotesize
\setlength{\tabcolsep}{4pt}
\centering
\caption{Accuracy on Irregular Character Trajectories}\label{tbl:charactertrajectory0}
\begin{tabular}{ccccc} 
\hline
\multirow{2}{*}{Model} & \multicolumn{3}{c}{Test Accuracy} & \multirow{2}{*}{\begin{tabular}[c]{@{}c@{}}Memory \\(MB)  \end{tabular}} \\ \cline{2-4}& 30\% dropped & 50\% dropped & 70\% dropped &                                                                           \\  \hline

GRU-$\Delta t$      & 0.936 ± 0.020     & 0.913 ± 0.021     & 0.904 ± 0.008     & 16.5  
\\
GRU-D               & 0.942 ± 0.021     & 0.902 ± 0.048     & 0.919 ± 0.017     & 17.8    
\\
GRU-ODE             & 0.926 ± 0.016     & 0.867 ± 0.039     & 0.899 ± 0.037     & 1.51   
\\
ODE-RNN             & 0.954 ± 0.006     & 0.960 ± 0.003     & 0.953 ± 0.006     & 15.5
\\
Latent-ODE          & 0.875 ± 0.027     & 0.869 ± 0.021     & 0.887 ± 0.059     & 181  
\\
Augmented-ODE       & 0.965 ± 0.014     & 0.953 ± 0.022     & 0.930 ± 0.029     & 186  
\\
ACE-NODE            & 0.876 ± 0.055     & 0.886 ± 0.025     & 0.910 ± 0.032     & 113   
\\
NCDE                & 0.987 ± 0.008     & 0.988 ± 0.002     & 0.986 ± 0.004     & 1.38                                                                    \\
ANCDE               & \textbf{0.992 ± 0.003}  & 0.989 ± 0.001     & 0.988± 0.002    & 2.02                                                            \\ \hline
\textbf{EXIT} & 0.991± 0.001 & \textbf{0.992 ± 0.002} & \textbf{0.992 ± 0.003}    & 3.79                                                    \\ \hline
\end{tabular}%
\end{table}

The time-series classification with PhysioNet Sepsis in Table~\ref{tbl:PhysioNet} is one of the most widely used benchmark experiments. 
Our method, EXIT, shows the best AUROC with significant differences from other baselines, when using the observation intensity, i.e., OI, and its GPU memory requirement is smaller than many other baselines. For this dataset, all CDE-based models show reasonable performance. When we do not use the observation intensity, i.e., No OI, GRU-D shows the best AUROC.

For the Speech Commands dataset, we summarize the results in Table~\ref{tbl:speech}. As summarized, all RNN/LSTM/GRU-based models are inferior to other differential equation-based models. We consider that this is because of the dataset characteristic. This dataset contains many audio signal samples and it is obvious that those physical phenomena can be well modeled as differential equations. Among many differential equation-based models, the two NCDE-based models, NCDE and EXIT, show the highest performance. However, EXIT significantly outperforms all others including NCDE. One more point is that our method requires much smaller GPU memory in comparison with many other baselines.

Tables~\ref{tbl:charactertrajectory1} and~\ref{tbl:charactertrajectory0} summarize the accuracy for Character Trajectories. To create an irregular time series environment, we randomly select 30\%, 50\%, and 70\% of the values for each sequence with a total length of $182$. Therefore, this is basically an irregular time-series classification. Many baselines show reasonable scores for the irregular setting. The three GRU-based models are specialized in processing irregular time-series and outperform some other ODE-based models. However, CDE-based models, including our EXIT, show the highest scores. Among them, especially, EXIT is clearly the best. Our method maintains an accuracy larger than 0.99 across all the dropping settings. For its regular setting, Table~\ref{tbl:charactertrajectory1} shows that our method shows the best score among the baselines. 

\subsection{Time Series Forecasting}

\paragraph{MuJoCo}
The Hopper model from the DeepMind control suite is used in this dataset~\cite{DBLP:journals/corr/abs-1801-00690}. 10,000 simulations of the Hopper model were used to generate it. This physics engine is used for research and development in domains like robotics and machine learning that demand precise simulations. This data is 14-dimensional, with 10,000 sequences of 100 regularly sampled time points each series. MSE is a metric that we employ.

\paragraph{Google Stock}
The Google Stock dataset includes the trading volumes of Google in conjunction with its high, low, open, close, and adjusted closing prices. Unlike the data used by ANCDE~\cite{jhin2021ancde}, we use the data from 2011 to 2021, the most recent and volatile years, to create a more challenging environment. CDEs were originally developed in the field of mathematical finance to predict various financial time-series values~\cite{Lyons1998}. Therefore, we expect that our EXIT shows the best appropriateness for this task. The goal is to predict, given past several days of time-series values, the high, low, open, close, and adjusted closing prices at the very next 10 days.

\paragraph{Experimental results}

\begin{table}[t]
\footnotesize
\setlength{\tabcolsep}{4pt}
\centering
\caption{MSE on Irregular MuJoCo}\label{tbl:mujoco0}
\begin{tabular}{ccccc} 
\hline
\multirow{2}{*}{Model} & \multicolumn{3}{c}{Test MSE} & \multirow{2}{*}{\begin{tabular}[c]{@{}c@{}}Memory\\(MB) \end{tabular}} \\ \cline{2-4}& 30\% dropped & 50\% dropped & 70\% dropped &                                                             \\  \hline
GRU-$\Delta t$  & 0.198 ± 0.036     & 0.193 ± 0.015     & 0.196 ± 0.028                     & 533   \\
GRU-D           & 0.608 ± 0.032     & 0.587 ± 0.039     & 0.579 ± 0.052                     & 569   \\
GRU-ODE         & 0.857 ± 0.015     & 0.852 ± 0.015     & 0.861 ± 0.015                     & 146   \\
ODE-RNN         & 0.274 ± 0.213     & 0.237 ± 0.110     & 0.267 ± 0.217                     & 115   \\
Latent-ODE      & 0.056 ± 0.001     & 0.055 ± 0.004     & 0.058 ± 0.003                     & 314   \\
Augmented-ODE   & 0.056 ± 0.004     & 0.057 ± 0.005     & 0.057 ± 0.005                     & 286   \\
ACE-NODE        & 0.053 ± 0.007     & 0.053 ± 0.005     & 0.052 ± 0.006                     & 423   \\
NCDE            & 0.027 ± 0.000     & 0.027 ± 0.001     & 0.026 ± 0.001                     & 52.1  \\
ANCDE           & 0.031 ± 0.002     & 0.029 ± 0.003     & 0.031 ± 0.002                     & 79.2  \\\hline
\textbf{EXIT}   & \textbf{0.025  ± 0.004} & \textbf{0.026 ± 0.000} & \textbf{0.026 ± 0.001} & 127   \\ \hline

\end{tabular}
\end{table}

\begin{table}[t]
\footnotesize
\setlength{\tabcolsep}{4pt}
\centering
\caption{MSE on Irregular Google Stock}\label{tbl:stock0}
\begin{tabular}{ccccc} 
\hline
\multirow{2}{*}{Model} & \multicolumn{3}{c}{Test MSE} & \multirow{2}{*}{\begin{tabular}[c]{@{}c@{}}Memory\\(MB) \end{tabular}} \\ \cline{2-4}& 30\% dropped & 50\% dropped & 70\% dropped &                                                             \\  \hline
GRU-$\Delta t$  & 0.145 ± 0.002     & 0.146 ± 0.001     & 0.145 ± 0.002                     & 13.2  \\
GRU-D           & 0.143 ± 0.002     & 0.145 ± 0.002     & 0.146 ± 0.002                     & 14.8   \\
GRU-ODE         & 0.064 ± 0.009     & 0.057 ± 0.003     & 0.059 ± 0.004                     & 23.7   \\
ODE-RNN         & 0.116 ± 0.018     & 0.145 ± 0.006     & 0.129 ± 0.011                     & 67.7   \\
Latent-ODE      & 0.052 ± 0.005     & 0.053 ± 0.001     & 0.054 ± 0.007                     & 20.9   \\
Augmented-ODE   & 0.045 ± 0.004     & 0.051 ± 0.005     & 0.057 ± 0.002                     & 31.6   \\
ACE-NODE        & 0.044 ± 0.002     & 0.053 ± 0.008     & 0.056 ± 0.003                     & 32.9   \\
NCDE            & 0.056 ± 0.015     & 0.054 ± 0.002     & 0.056 ± 0.007                     & 52.8  \\
ANCDE           & 0.048 ± 0.012     & 0.047 ± 0.001     & 0.049 ± 0.004                     & 10.2  \\\hline
\textbf{EXIT}   & \textbf{0.042  ± 0.001} & \textbf{0.045 ± 0.001} & \textbf{0.046 ± 0.002} & 29.4   \\ \hline

\end{tabular}
\end{table}

\begin{table}[t]
\begin{minipage}{.48\linewidth}
\setlength{\tabcolsep}{1pt}
\scriptsize
\centering
\caption{MSE on Regular MuJoCo}\label{tbl:mujuco1}
\begin{tabular}{ccc}
\hline
Model & Test MSE & Memory \\ 
\hline
RNN             & 0.063 ± 0.001             & 409 \\
LSTM            & 0.064 ± 0.001             & 411 \\
GRU             & 0.063 ± 0.000             & 439 \\
GRU-$\Delta t$  & 0.223 ± 0.020             & 533 \\
GRU-D           & 0.578 ± 0.042             & 569 \\
GRU-ODE         & 0.856 ± 0.016             & 146 \\
ODE-RNN         & 0.328 ± 0.225             & 115 \\
Latent-ODE      & 0.029 ± 0.011             & 314 \\
Augmented-ODE   & 0.055 ± 0.004             & 286 \\
ACE-NODE        & 0.039 ± 0.003             & 423  \\
NCDE            & 0.028 ± 0.002             & 52.1 \\
ANCDE           & 0.029 ± 0.003             & 79.2 \\
\hline
\textbf{EXIT}   & \textbf{0.026 ± 0.000}    & 127  \\
\hline
\end{tabular}
\end{minipage} \hfill
\begin{minipage}{.48\linewidth}
\setlength{\tabcolsep}{1pt}
\scriptsize
\centering
\caption{MSE on Regular Google Stock}\label{tbl:stock1}
\begin{tabular}{ccc}
\hline
Model & Test MSE & Memory \\ 
\hline
RNN             & 0.058 ± 0.018             & 30.6 \\
LSTM            & 0.075 ± 0.045             & 25.8 \\
GRU             & 0.073 ± 0.037             & 27.6 \\
GRU-$\Delta t$  & 0.126 ± 0.002             & 13.2 \\
GRU-D           & 0.140 ± 0.004             & 14.8 \\
GRU-ODE         & 0.068 ± 0.016             & 23.7 \\
ODE-RNN         & 0.111 ± 0.044             & 67.7 \\
Latent-ODE      & 0.053 ± 0.011             & 20.9 \\
Augmented-ODE   & 0.053 ± 0.002             & 31.6 \\
ACE-NODE        & 0.049 ± 0.003             & 32.9 \\
NCDE            & 0.057 ± 0.062             & 52.8 \\
ANCDE           & 0.046 ± 0.002             & 10.2 \\
\hline
\textbf{EXIT}   & \textbf{0.042 ± 0.002}    & 29.4 \\
\hline
\end{tabular}
\end{minipage}
\end{table}

For the MuJoCo dataset, we drop randomly 30\%, 50\%, and 70\% of values in each sequence to create challenging environments, i.e., irregular time-series forecasting. In Table~\ref{tbl:mujoco0}, our method, EXIT, clearly shows the best MSE for all dropping ratios. One outstanding point in our research is that the MSE is not greatly influenced by the dropping ratio but maintains its small error across all the dropping ratios. Also, Table~\ref{tbl:mujuco1} shows the best MSE for the regular setting of MuJoCo.

The irregular time-series forecasting results of Google Stock are summarized in Table~\ref{tbl:stock0}. Among ODE-based models, ACE-NODE shows the smallest standard deviation and the mean MSE values. Among CDE-based models, NCDE and ANCDE shows reasonably small standard deviation and small MSE values. However, EXIT shows the smallest mean MSE and standard deviation among all methods for this dataset. Also, Table~\ref{tbl:stock1} shows the best MSE for the regular setting of Google Stock.

\subsection{Learned Integral Time Duration}
\begin{table}[t]
\setlength{\tabcolsep}{3pt}
\footnotesize
\centering
\caption{The learned time $\tau_{end}$ is smaller than the terminal time $T$ of data}\label{tbl:tau_interp}

\begin{tabular}{cccc}
\hline
DataSet                  &      $\tau_{start}$  &       $\tau_{end}$    &   $T$    \\ \hline
PhysioNet Sepsis (OI)    &      5.0851          &       65.9149         &   71     \\
PhysioNet Sepsis (No OI) &      1.1980          &       69.8020         &   71     \\
Character Trajectories   &      3.0538          &       177.9465        &   181    \\
MuJoCo                   &      0.1152          &       48.8848         &   49     \\\hline
\end{tabular}
\end{table}

\begin{table}[t]
\setlength{\tabcolsep}{3pt}
\footnotesize
\centering
\caption{The learned time $\tau_{end}$ is beyond the terminal time $T$ of data}\label{tbl:tau_extrap}

\begin{tabular}{cccc}
\hline
DataSet            &      $\tau_{start}$  &   $T$            &   $\tau_{end}$  \\ \hline
Speech Commands    &      0               &   160            &   161.3584    \\
Google Stock       &      0.80313         &   49             &   49.49       \\ \hline
\end{tabular}
\end{table}

Tables~\ref{tbl:tau_interp} and~\ref{tbl:tau_extrap} show the learned integral time duration. In PhysioNet Sepsis, Character trajectories and MuJoCo, it is learned that $0 < \tau_{start} < \tau_{end} < T$. In the case of PhysioNet Sepsis with OI, the start time $\tau_{start}$ increases to 5.0851 and the terminal time $\tau_{end}$ is learned to 65.9149 which is smaller than $T$. In the case of Character Trajectories, $\tau_{start}$ increases from 0 to $3.0538$, and $\tau_{end}$ decreases from 181 to 177.9465. 

For Speech Commands and Google Stock, as shown in Table~\ref{tbl:tau_extrap}, the learned pattern is different from the previous four cases in Table~\ref{tbl:tau_interp}. In the case of Speech Commands, $\tau_{end}$ increases to 161.3584, which is beyond the terminal time of data, and in Google Stock, a similar pattern is observed.

\subsection{Ablation/Sensitivity Studies and Others}

\paragraph{Sensitivity to the time learning rate $\lambda_{\tau}$}

\begin{figure}[!t]
    \centering
    \subfigure[Sensitivity to $\lambda_{\tau}$ in Speech Commands]{\includegraphics[width=0.49\columnwidth]{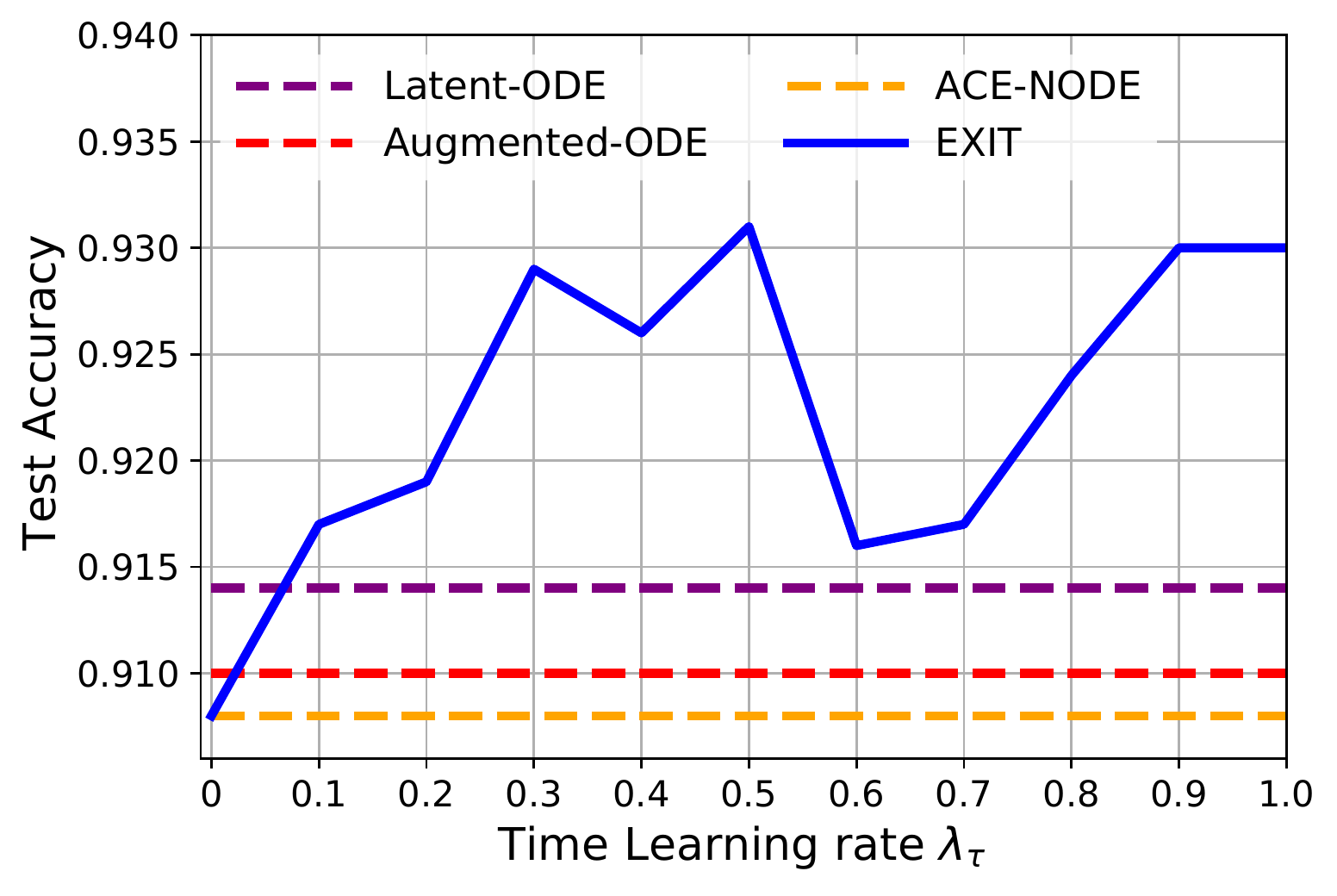}}\hfill
    \subfigure[MSE by the output size in Google Stock]{\includegraphics[width=0.49\columnwidth]{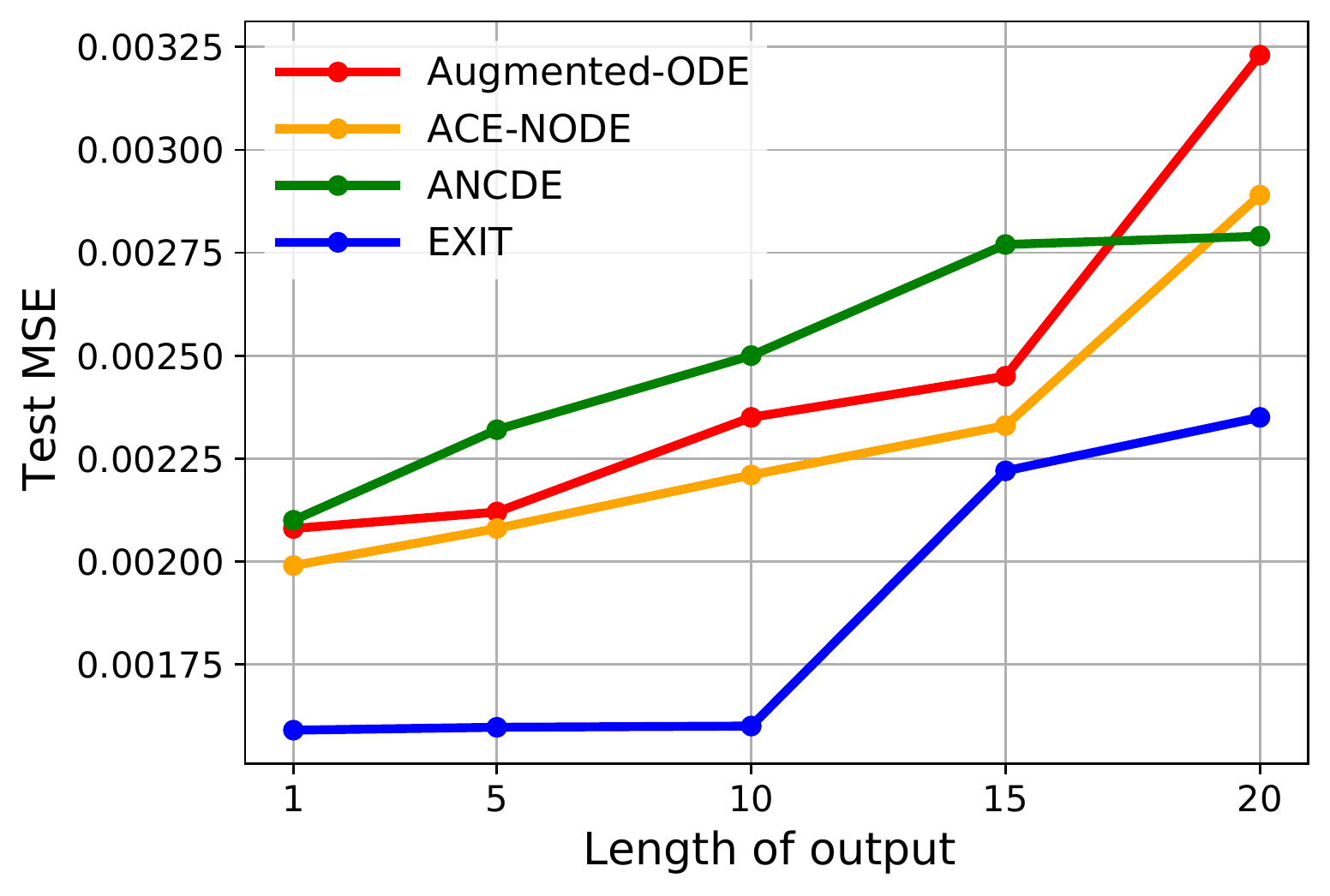}}
        \subfigure[PhysioNet Sepsis]{\includegraphics[width=0.49\columnwidth]{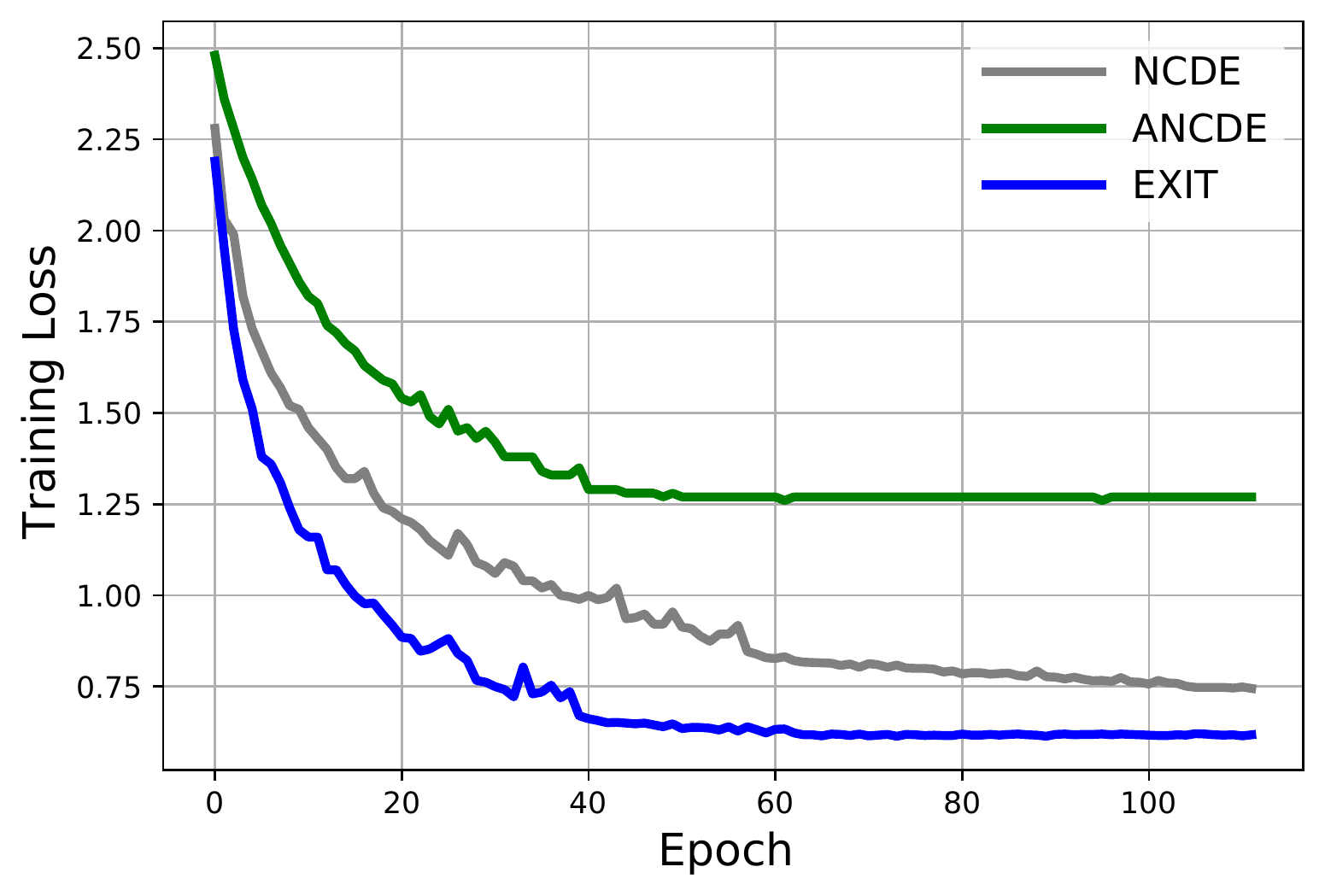}}\hfill
    \subfigure[Speech Commands]{\includegraphics[width=0.49\columnwidth]{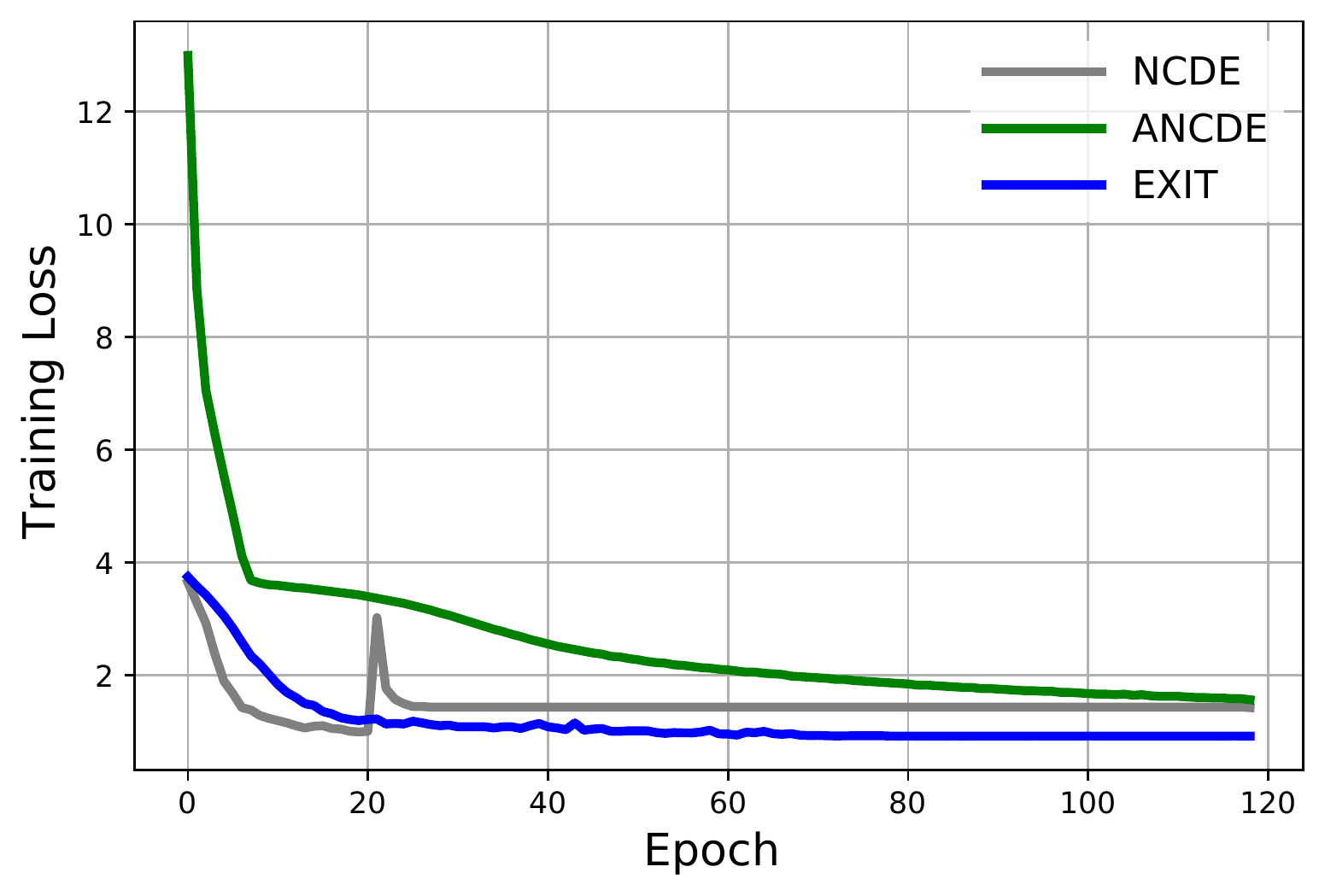}}
    \caption{Sensitivity analysis, ablation study, and training loss curves. More figures are in Appendix.}
    \label{fig:vis_ablation}
\end{figure}
Fig.~\ref{fig:vis_ablation} (a) shows the sensitivity curve w.r.t. the time learning rate. The model accuracy varies a lot depending on the time learning rate, but it can be seen that most of them are better than the top-3 baselines of Table~\ref{tbl:speech}. As the time learning rate increases, $\tau_{start}$ and $\tau_{end}$ change a lot, and it shows the best accuracy when the time learning rate is large enough. The same pattern can also be observed for other datasets.

\paragraph{Ablation study on the output sequence length}
Fig.~\ref{fig:vis_ablation} (b) shows the ablation study curve w.r.t. the output sequence length. We also compare our model with the top-3 baselines of Table~\ref{tbl:stock0}. After fixing the input length to 50 days, we variate the output length from 1 to 20 days. Our method, EXIT, consistently outperforms others.




\paragraph{Training loss curves} In Figs.~\ref{fig:vis_ablation} (c) and (d), we compare the training loss curves of various NCDE-based methods. Among them, our method shows the fastest training speed and better minimizes the loss. In Speech Commands, moreover, NCDE does not show a reliable training curve whereas our method, EXIT, shows a very reliable curve.

\section{Conclusions and Future Work}
For the past couple of years, differential equation-based methods have proliferated for time-series classification and forecasting. Many advanced methods have been proposed ever since the first introduction to NODEs. Among them, NCDEs are one of the most influential breakthroughs. A core part of NCDEs is how to interpolate discrete time-series samples and generate continuous paths. In this paper, we presented how to interpolate and extrapolate a latent path with our proposed encoder-decoder architecture, and as a result, NCDEs' model performance can be significantly improved. Based on 5 real-world datasets and 12 baselines, various experiments were conducted, ranging from regular/irregular classification to regular/irregular forecasting. Our method, EXIT, performs the best in most cases.

We explored the possibility of improving NCDEs by learning a latent path and controlling the integral time duration on the latent path. However, NCDEs are a recent breakthrough and they are not fully studied yet. We hope that our research encourages much follow-up research on improving NCDEs with novel methods.

\section*{Acknowledgement}
Noseong Park is the corresponding author. This work was supported by the Yonsei University Research Fund of 2021, and the Institute of Information \& Communications Technology Planning \& Evaluation (IITP) grant funded by the Korean government (MSIT) (No. 2020-0-01361, Artificial Intelligence Graduate School Program (Yonsei University), and No. 2021-0-00155, Context and Activity Analysis-based Solution for Safe Childcare).

\bibliography{ref}
\bibliographystyle{ACM-Reference-Format}

\clearpage

\appendix
\section{SW/HW Environments \& Best Hyperparameters}
Our software and hardware environments are as follows: \textsc{Ubuntu} 18.04 LTS, \textsc{Python} 3.7.10, \textsc{Pytorch} 1.8.1, \textsc{CUDA} 11.4, and \textsc{NVIDIA} Driver 470.42.01, i9 CPU, and \textsc{NVIDIA RTX 8000}.

We summarize the best hyperparameter set for our method in each dataset in Table~\ref{tbl:besthyper}. The best CDE/ODE functions are listed in Tables~\ref{tbl:mujocok} to~\ref{tbl:googlegf}.

\begin{table}[t]
\setlength{\tabcolsep}{4pt}
\footnotesize
\centering
\caption{Best Hyperparameter for all dataset}\label{tbl:besthyper}
\begin{tabular}{cccccc}
\hline
\multirow{2}{*}{Hyperparameter} & \multicolumn{5}{c}{Dataset} \\ \cline{2-6} 
& \begin{tabular}[c]{@{}c@{}}PhysioNet \\ Sepsis\end{tabular} & \begin{tabular}[c]{@{}c@{}}Character \\ Trajectories\end{tabular} & \begin{tabular}[c]{@{}c@{}}Speech \\ Commands\end{tabular} & MuJoCo & Stock Google \\ \hline
$c_{wd}$          & 1.0       & 0.1         & 0.1       & 1.0       & 1.0   \\
$\lambda$         & 0.005     & 0.0005      & 0.0005    & 0.001     & 0.01  \\
$\lambda_{\tau}$  & 1.0       & 0.1         & 1.0       & 1.0       & 0.001 \\
$c_{kr}$          & 1e-3      & 1e-3        & 1e-3      & 0         & 1e-5  \\
epoch             & 70        & 130         & 100       & 300       & 70    \\ \hline
\end{tabular}
\end{table}

\begin{table}[t]
\scriptsize
\setlength{\tabcolsep}{4pt}
\caption{The best architecture of the CDE function $k$  for MuJoCo}\label{tbl:mujocok}
\begin{tabular}{cccc}
\hline
Design                  & Layer & Input                 & Output            \\ \hline
\texttt{FC}             & 1     & $1024  \times $100    & $1024 \times $40  \\
$\rho$(\texttt{FC})     & 2     & $1024  \times $40     & $1024 \times $40  \\
$\rho$(\texttt{FC})     & 3     & $1024  \times $40     & $1024 \times $40  \\
$\rho$(\texttt{FC})     & 4     & $1024  \times $40     & $1024 \times $40  \\
$\xi$(\texttt{FC})      & 5     & $1024  \times $40     & $1024 \times $1400 \\ \hline
\end{tabular}
\end{table}

\begin{table}[t]
\scriptsize
\setlength{\tabcolsep}{4pt}
\caption{The best architecture of the CDE function $g$ for MuJoCo}\label{tbl:mujocog}
\begin{tabular}{cccc}
\hline
Design                  & Layer & Input                 & Output            \\ \hline
\texttt{FC}             & 1     & $1024  \times  $100   & $1024 \times $40  \\
$\rho$(\texttt{FC})     & 2     & $1024  \times  $40    & $1024 \times $40  \\
$\rho$(\texttt{FC})     & 3     & $1024  \times  $40    & $1024 \times $40  \\
$\rho$(\texttt{FC})     & 4     & $1024  \times  $40    & $1024 \times $40  \\
$\xi$(\texttt{FC})      & 5     & $1024  \times  $40    & $1024 \times $100 \\ \hline
\end{tabular}
\end{table}

\begin{table}[t]
\scriptsize
\setlength{\tabcolsep}{4pt}
\caption{The best architecture of the ODE function $f$ for MuJoCo}\label{tbl:mujocof}
\begin{tabular}{cccc}
\hline
Design                          & Layer & Input                 & Output            \\ \hline
\texttt{FC}                     & 1     & $1024 \times $100     & $1024 \times $40  \\
$\varepsilon$(\texttt{FC})      & 2     & $1024 \times $40      & $1024 \times $40  \\
$\varepsilon$(\texttt{FC})      & 3     & $1024 \times $40      & $1024 \times $40  \\
$\varepsilon$(\texttt{FC})      & 4     & $1024 \times $40      & $1024 \times $40  \\
$\xi$(\texttt{FC})              & 5     & $1024 \times $40      & $1024 \times $100 \\ \hline
\end{tabular}
\end{table}

\begin{table}[t]
\scriptsize
\setlength{\tabcolsep}{4pt}
\caption{The best architecture of the CDE function $k$ for Google Stock}\label{tbl:googlek}
\begin{tabular}{cccc}
\hline
Design                  & Layer & Input                 & Output            \\ \hline

$\sigma$(\texttt{FC})   & 1  & $256 \times $80    & $256 \times $80  \\
$\sigma$(\texttt{FC})   & 2  & $256 \times $80    & $256 \times $80  \\
$\xi$(\texttt{FC})   & 3  & $256 \times $80    & $256 \times $80  \\
\hline
\end{tabular}
\end{table}

\begin{table}[t]
\scriptsize
\setlength{\tabcolsep}{4pt}
\caption{The best architecture of the CDE function $g$ and the ODE function $f$ for Google Stock}\label{tbl:googlegf}
\begin{tabular}{cccc}
\hline
Design                  & Layer & Input                 & Output            \\ \hline
\texttt{FC}             & 1  & $256 \times $100  & $256 \times $40  \\
$\varepsilon$(\texttt{FC})     & 2  & $256 \times $40   & $256 \times $40  \\
$\varepsilon$(\texttt{FC})     & 3  & $256 \times $40   & $256 \times $40  \\
$\varepsilon$(\texttt{FC})     & 4  & $256 \times $40   & $256 \times $40  \\
$\xi$(\texttt{FC})             & 5  & $256 \times $40   & $256 \times $1400 \\ \hline
\end{tabular}
\end{table}

For reproducibility, we also report the best hyperparameters of other baselines as follows:
\begin{enumerate}
\item In PhysioNet Sepsis, we train for 100 epochs with a batch size of 1,024, and stopped early if the train loss doesn't decrease for 50 epochs. A hidden vector size in $\{50,60,70\}$ and a learning rate in $\{\num{1.0e-4}, \num{5.0e-4}, \num{1.0e-3}, \num{5.0e-3}\}$ are used.

\item In Character Trajectories, we train for 150 epochs with a batch size of 32, and stopped early if the train loss doesn't decrease for 50 epochs. A hidden size in $\{10, 20, 40, 50, 60\}$ and a learning rate in $\{\num{1.0e-4}, \num{5.0e-4}, \num{1.0e-3}, \num{5.0e-3}\}$ are used. For GRU, LSTM, and RNN, we use a hidden vector size of 40.

\item In Speech Commands, we train for 150 epochs with a batch size of 1,024, and stopped early if the train loss doesn't decrease for 50 epochs. We use a hidden vector size of $\{60,90,120\}$, and a learning rate of $\{\num{1.0e-6}, \num{1.0e-5}, \num{1.0e-4}\}$. For GRU, LSTM, and RNN, we use a hidden vector size of 160.

\item In MuJoCo, we train for 500 epochs with a batch size of 1,024, and stopped early if the train loss doesn't decrease for 50 epochs. We use a hidden vector size of $\{50,60,70,75\}$, and a learning rate of $\{\num{1.0e-5},\num{1.0e-4}, \num{1.0e-3}\}$. For GRU, LSTM, and RNN, we use a hidden vector size of 180.

\item In Google Stock, we train for 300 epochs with a batch size of 256, and stopped early if the train loss doesn't decrease for 50 epochs. We use a hidden vector size of $\{40,50,60,80\}$, and a learning rate of $\{\num{1.0e-5}, \num{1.0e-4},\num{1.0e-3}\}$. For GRU, LSTM, and RNN, we use a hidden vector size of 20.
\end{enumerate}

\section{Additional Experimental Results}
\begin{figure}[H]
    \centering
    \includegraphics[width=0.8\columnwidth]{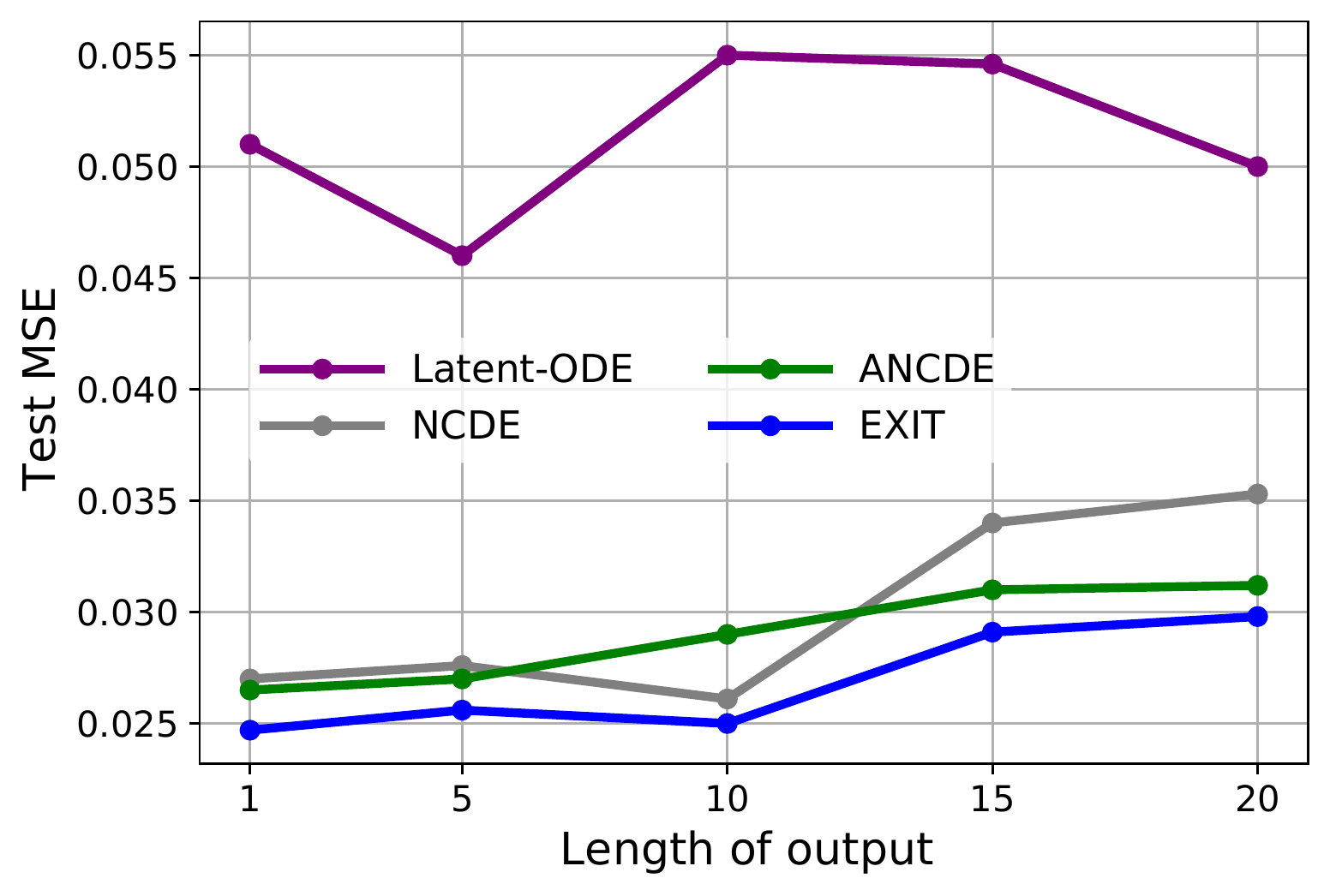}
    \caption{Error by the output length in MuJoCo}
    \label{fig:mujoco_seq}
\end{figure}

\begin{figure}[H]
    \centering
    \subfigure[MuJoCo]{\includegraphics[width=0.75\columnwidth]{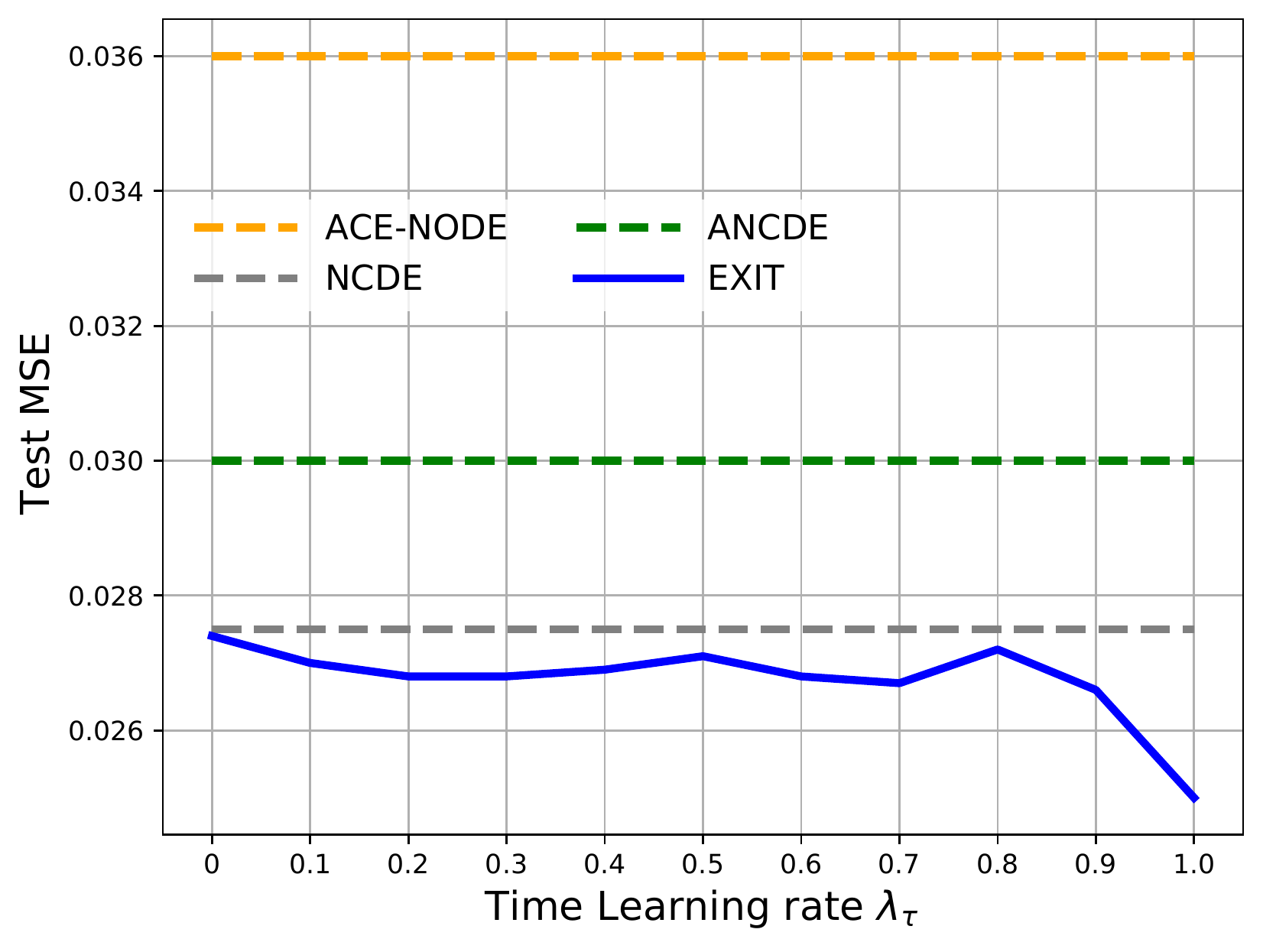}}\hfill
    \subfigure[Google Stock]{\includegraphics[width=0.75\columnwidth]{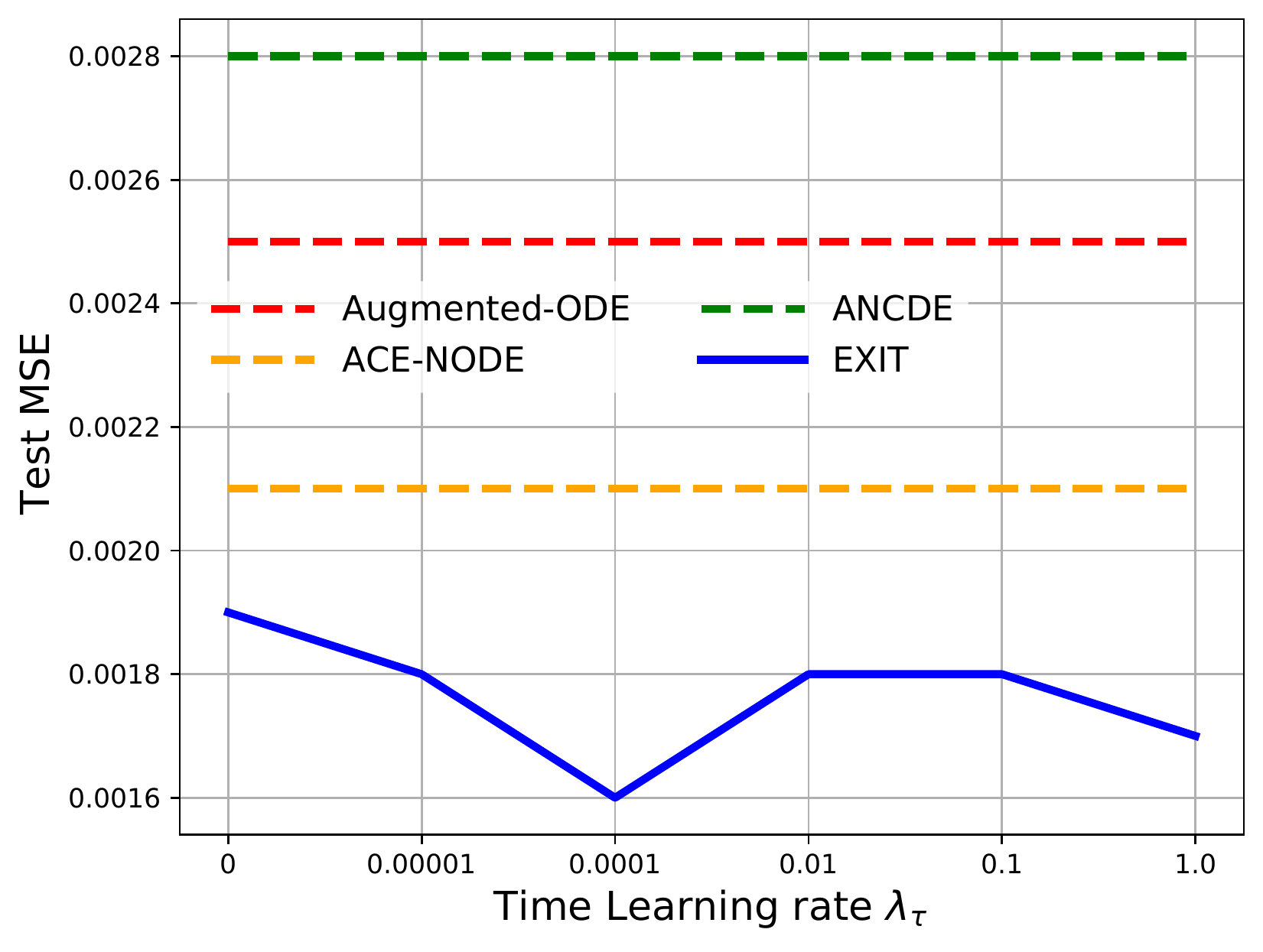}}\\
    
    \subfigure[PhysioNet Sepsis]{\includegraphics[width=0.75\columnwidth]{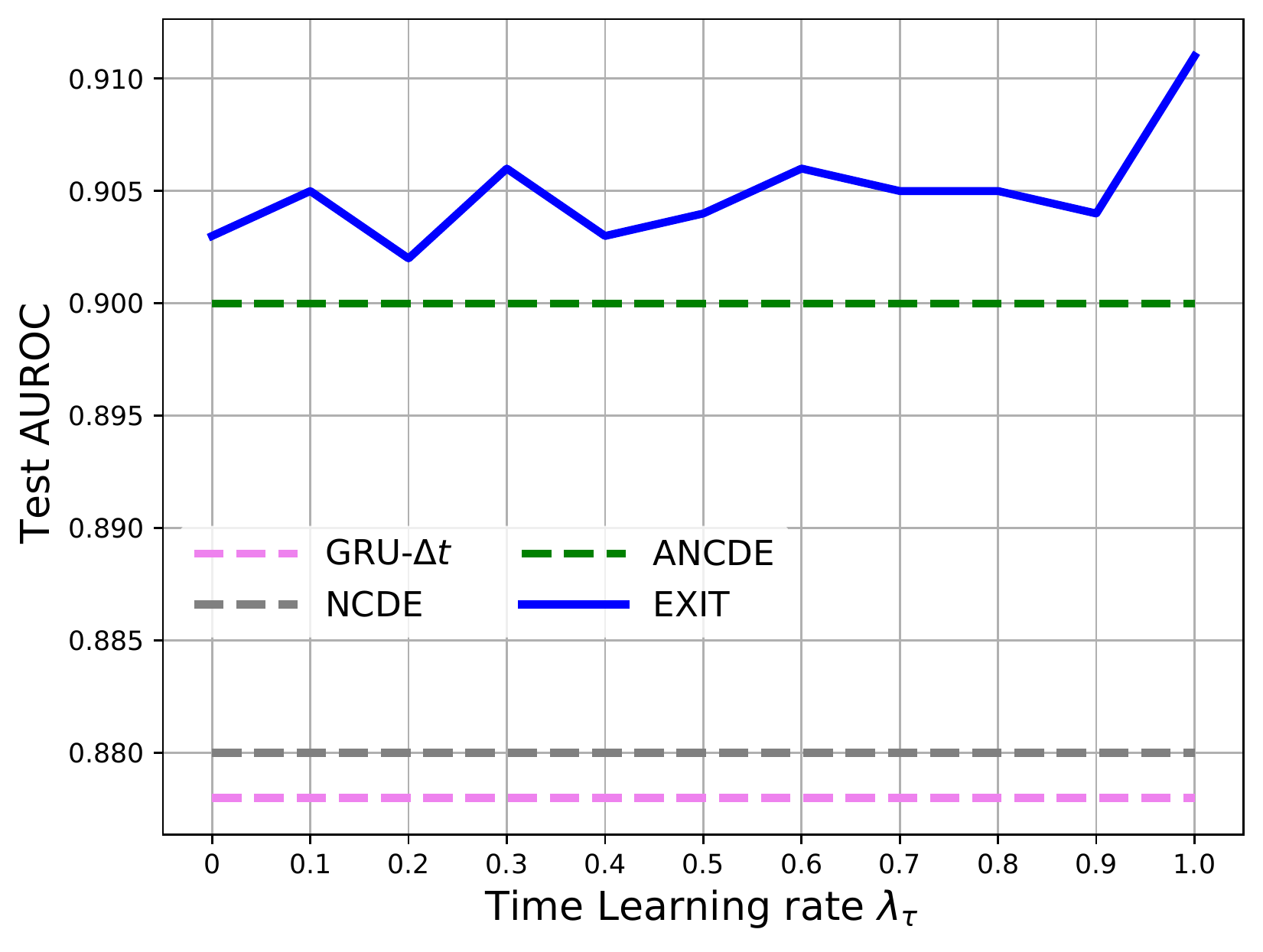}}\hfill 
    \subfigure[Character Trajectories]{\includegraphics[width=0.75\columnwidth]{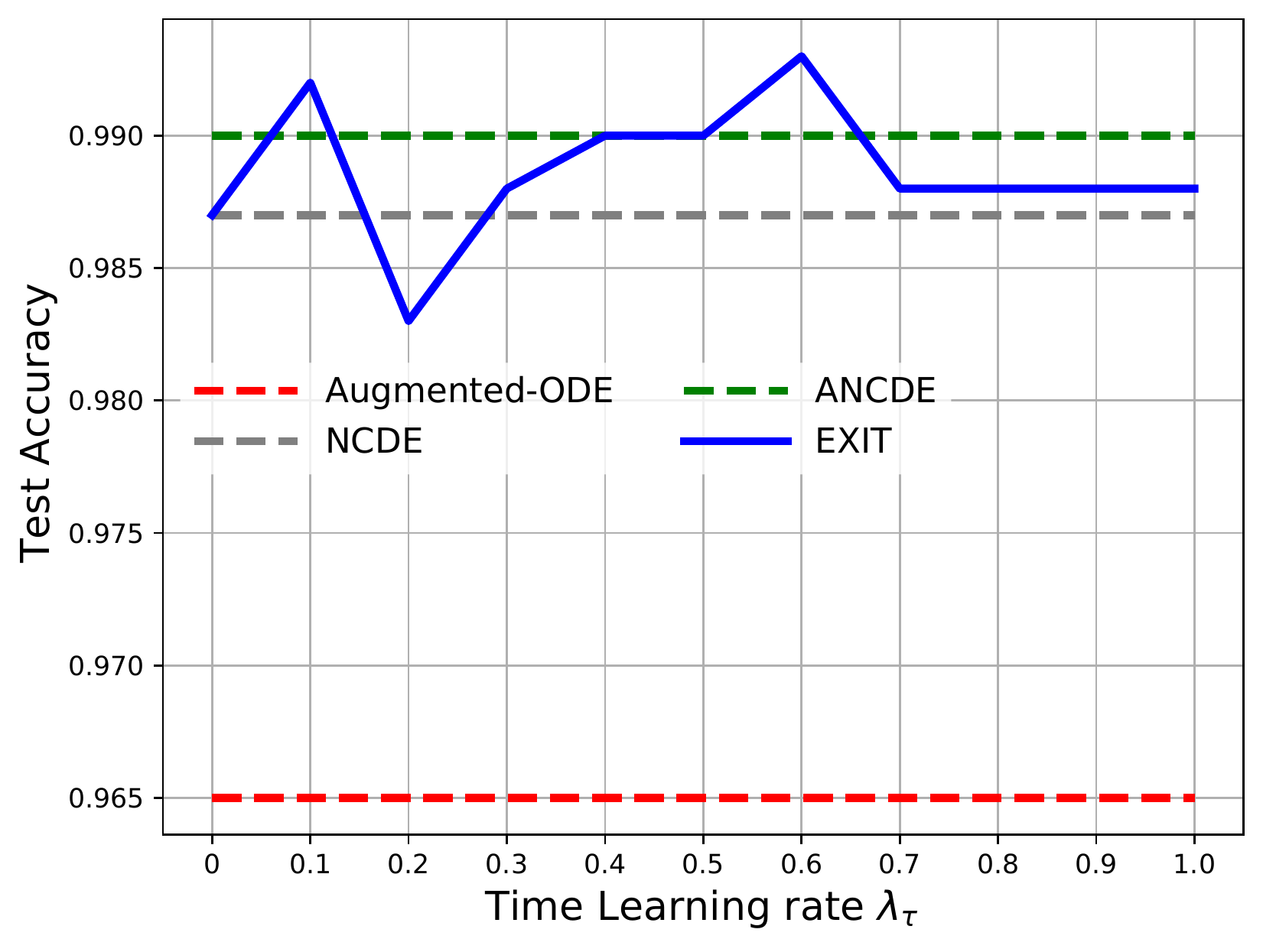}}
    \caption{Sensitivity to the time learning rate $\lambda_{\tau}$}
    \label{fig:vis_ablation4}
\end{figure}

We also report additional ablation and sensitivity study results in Figs.~\ref{fig:mujoco_seq} and~\ref{fig:vis_ablation4}. 

\section{Ablation study}
\paragraph{Ablation study after fixing $\tau_{start}$ and $\tau_{end}$}

\begin{table}[t]
\setlength{\tabcolsep}{4pt}
\footnotesize
\centering
\caption{Test on fixed time}\label{tbl:abl1}
\begin{tabular}{cccc}
\hline
Dataset & Fixed-EXIT & Terminal-EXIT & \textbf{EXIT}\\ 
\hline
PhysioNet Sepsis (AUROC)           & 0.903 & 0.905 & \textbf{0.913} \\
Speech Commands (Accuracy)         & 0.860 & 0.926 & \textbf{0.930} \\
Character Trajectories (Accuracy)  & 0.987 & 0.990 & \textbf{0.992} \\
MuJoCo (MSE)                       & 0.028 & 0.027 & \textbf{0.025} \\
Google Stock (MSE)                 & 0.019 & 0.018 & \textbf{0.016} \\
\hline
\end{tabular}
\end{table}

Table~\ref{tbl:abl1} shows the ablation study on the integral time duration. Fixed-EXIT in the table is a case where both the start and end time values are fixed to 0 and $T$, respectively. Terminal-EXIT is a case where the start time is fixed to 0 and only the end integral time is learned. Terminal-EXIT shows better results than Fixed-EXIT, but EXIT, which learns both the start and end integral time, shows the best results.

\end{document}